\documentclass{article}

 \usepackage[preprint]{neurips_2026}

\usepackage[utf8]{inputenc} 
\usepackage[T1]{fontenc}    

\usepackage{hyperref}       
\usepackage{url}            
\usepackage{booktabs}       
\usepackage{amsfonts}       
\usepackage{nicefrac}       
\usepackage{microtype}      
\usepackage{xcolor}         

\usepackage{natbib}
\usepackage{graphicx}
\usepackage{amsmath}
\usepackage{amssymb}
\usepackage{array}
\usepackage{multirow}
\usepackage{color}
\usepackage{colortbl}
\usepackage{framed}
\usepackage{bm}
\usepackage{xspace}
\usepackage{enumitem}
\usepackage{xparse}
\usepackage{algorithm}
\usepackage{listings}
\usepackage{url}
\usepackage{mathtools}
\usepackage{pifont}
\usepackage{mathtools}
\usepackage{lipsum}
\usepackage{adjustbox}
\usepackage{bbm}
\usepackage{makecell}
\usepackage{soul} 
\usepackage{circledsteps}

\newcommand{\cmark}{\ding{51}}

\definecolor{Gray}{gray}{0.93}
\definecolor{tablegray}{RGB}{122, 122, 122}
\definecolor{lightgray}{gray}{0.92}
\definecolor{blond}{rgb}{0.98, 0.94, 0.75}
\definecolor{TitleColor}{gray}{0.95}
\definecolor{LightCyan}{rgb}{0.88,0.95,1}
\definecolor{OurColor}{RGB}{252, 230, 197}

\definecolor{blond}{rgb}{0.98, 0.94, 0.75}

\def \eg {\emph{e.g.}}

\newcommand{\tit}[1]{\noindent\textbf{#1.}}

\definecolor{myPink}{RGB}{247, 219, 231}  
\definecolor{myGreen}{RGB}{227, 241, 217}   
\definecolor{myBlue}{RGB}{187, 230, 240}    
\definecolor{myYellow}{RGB}{255, 242, 204}    
\definecolor{myOrange}{RGB}{255, 229, 204}    
\definecolor{myPurple}{RGB}{224, 208, 250}    
\definecolor{githubOrange}{RGB}{235, 137, 0} 

\newcommand{\hlpink}[1]{\sethlcolor{myPink}\hl{#1}}

\newcommand{\hlblue}[1]{\sethlcolor{myBlue}\hl{#1}}

\newcommand{\hlorange}[1]{\sethlcolor{myOrange}\hl{#1}}

\title{Few Channels Draw The Whole Picture: Revealing Massive Activations in Diffusion Transformers}

\author{%
  Evelyn Turri*\textsuperscript{1}
  ~
  Davide Bucciarelli*\textsuperscript{1,2}
  \\
  \textbf{Sara Sarto}\textsuperscript{1}
  ~
  \textbf{Lorenzo Baraldi}\textsuperscript{1}
  ~
  \textbf{Marcella Cornia}\textsuperscript{1}
  \\
  \textsuperscript{1}University of Modena and Reggio Emilia, Italy \quad \textsuperscript{2}University of Pisa, Italy\\
  \texttt{\{name\}.\{surname\}@unimore.it} \quad \small{*Equal contribution}\\
\\
{\tt\small \href{https://aimagelab.github.io/MAs-DiT/}{\textcolor{githubOrange}{aimagelab.github.io/MAs-DiT}}} 
}

\begin{document}


\maketitle
\vspace{-0.5cm}
\begin{abstract}
Diffusion Transformers (DiTs) and related flow-based architectures are now among the strongest text-to-image generators, yet the internal mechanisms through which prompts shape image semantics remain poorly understood. In this work, we study \textit{massive activations}: a small subset of hidden-state channels whose responses are consistently much larger than the rest. We show that, despite their sparsity, these few channels effectively \textit{draw the whole picture}, in three complementary senses. First, they are functionally critical: a controlled disruption probe that zeroes the massive channels causes a sharp collapse in generation quality, while disrupting an equally-sized set of low-statistic channels has marginal effect. Second, they are spatially organized: restricting image-stream tokens to massive channels and clustering them yields coherent partitions that closely align with the main subject and salient regions, exposing a structured spatial code hidden inside an apparently outlier-like subspace. Third, they are transferable: transporting massive activations from one prompt-conditioned trajectory into another, shifts the final image toward the source prompt while preserving substantial content from the target, producing localized semantic interpolation rather than unstructured pixel blending. We exploit this property in two use cases: \textit{text-conditioned} and \textit{image-conditioned semantic transport}, where massive activations transport enables prompt interpolation and subject-driven generation without any additional training. Together, these results recast massive activations not as activation anomalies, but as a sparse prompt-conditioned carrier subspace that organizes and controls semantic information in modern DiT models.
\end{abstract}

\section{Introduction}
\label{sec:introduction}

Text-to-image generation has shifted from convolutional denoisers to Transformer-based architectures~\cite{vaswani2017attention}. Diffusion Transformers (DiTs)~\cite{bao2023all,flux2024,peebles2023scalable} replace the U-Net~\cite{dhariwal2021diffusion,ho2020denoising,rombach2022high} with a Transformer backbone, enabling scalable, high-quality image synthesis. Recent models such as FLUX~\cite{flux2024,flux-2-2025}, SANA~\cite{xie2025sana}, and Qwen-Image~\cite{wu2025qwen} highlight the flexibility of this paradigm. However, how these models internally represent and propagate semantic information from prompts is still poorly understood.
A growing body of work shows that diffusion models contain rich internal semantic structure~\cite{bucciarelli2026tiny,hedlin2023unsupervised,tang-etal-2023-daam,zhang2023tale}. Prior approaches leverage attention maps for localization and editing~\cite{shin2025exploring,10654973}, while more recent studies suggest that DiTs exhibit emergent interpretability without supervision~\cite{erelattention, helbling2025conceptattention}. In parallel, recent analyses have identified \emph{massive activations} (MAs), defined as a small subset of hidden channels with exceptionally large magnitude~\cite{ganunleashing, gan2026massive}. However, their functional role remains unclear. Some works treat them as outliers that degrade feature quality, particularly for dense visual correspondence tasks~\cite{ganunleashing}, while others argue that they primarily contribute to local detail synthesis and have limited influence on global semantics~\cite{sun2024massive}. This raises a fundamental question: \emph{do MAs merely reflect outliers, or do they control meaningful semantic information?}

\begin{figure}[t]
    \centering
    \includegraphics[width=0.98\linewidth]{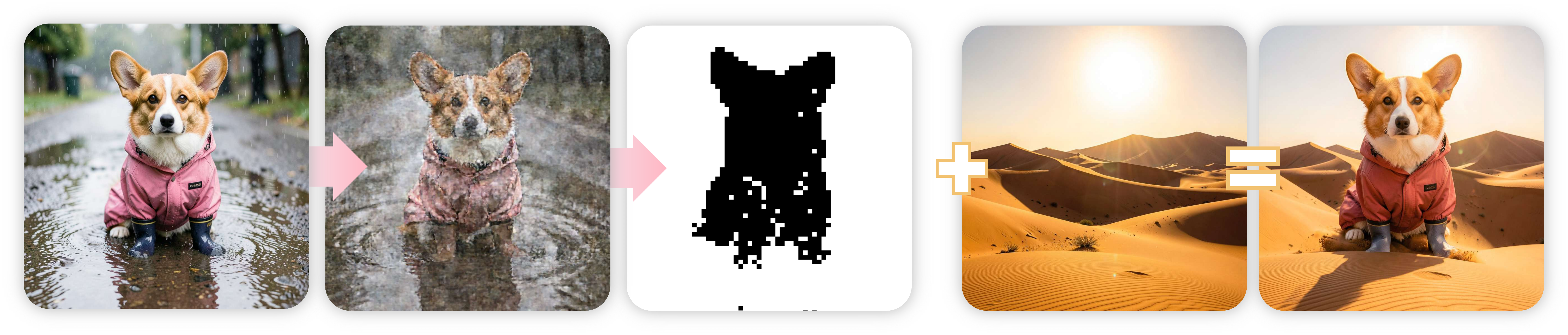}
    \vspace{-0.15cm}
    \caption{Overview of our findings. (\hlpink{Left}) Disrupting the top-$k$ massive activation channels severely degrades generation quality, revealing their functional importance. (\hlorange{Center}) These channels exhibit structured spatial organization, which we capture as a mask. (\hlblue{Right}) Transplanting the top-$k$ activations within this mask from a source to a target generation enables localized semantic transfer, producing coherent compositions rather than pixel-level blending.}
    \label{fig:intro}
    \vspace{-0.3cm}
\end{figure}

In this work, we examine the role of MAs in generation, identifying their concrete effects on semantic structure and controllability in the output.
Rather than studying them only through representation quality or downstream tasks, we investigate whether they constitute a sparse subspace that mediates both semantic organization and controllability in diffusion models.

Our first finding is \hlpink{structural}. While prior works show that class-conditioning vectors are sparse and exhibit properties of MAs~\cite{phamhidden}, we demonstrate that this phenomenon is not limited to conditioning signals. Through a channel disruption analysis, we show the role of persistent MAs across layers, denoising steps, and both streams in MMDiT architectures.

Our second finding is \hlorange{spatial}. Restricting representations to the top-$k$ massive channels reveals structured activation patterns that consistently align with semantically meaningful regions, such as foreground objects and background. Despite their sparsity in channel space, these activations induce a compact representation in which semantic regions become more separable.

Our third finding is \hlblue{intervention-based}. 
By transplanting top-$k$ MAs from one prompt-conditioned trajectory into another, we achieve controlled and localized semantic shifts.
The resulting images move toward the source prompt while preserving substantial content from the target, yielding coherent semantic compositions rather than pixel-level mixtures.

Building on this property, we demonstrate two practical use cases: \Circled{1} \emph{text-conditioned semantic transport}, where MAs transplanted across prompt-conditioned trajectories yield localized prompt combination, and \Circled{2} \emph{image-conditioned semantic transport}, where subject-specific activations extracted from a reference image are re-injected into new generations. Together, these results recast MAs as a \emph{sparse semantic transport subspace}: privileged coordinates that stabilize prompt-conditioned information across denoising steps, localize it in space, and transmit it to the final image, reconciling prior views of these channels as mere outliers.

Our contributions are as follows: (i) We extend prior work with a channel-disruption probe tailored to MMDiT architectures, analyzing MAs across five state-of-the-art DiTs~\cite{flux2024,flux-2-2025,wu2025qwen,xie2025sana}. (ii) We show that massive image-stream channels induce semantically meaningful spatial structure, validated segmenting GenAI-Bench~\cite{li2024genai} generated images. (iii) We introduce a masked activation-transport mechanism that, without training, enables controllable semantic composition on GenAI-Bench and matches dedicated editing backbones on DreamBench++~\cite{peng2024dreambench} personalization benchmark.
\section{Related Work}
\label{sec:related}
\tit{Modern Flow-Matching DiTs for Text-to-Image}
\label{sec:related-mmdit}
Recent advances in text-to-image generation have shifted from U-Net-based diffusion backbones~\cite{ho2020denoising,rombach2022high} to Transformer-based architectures, namely Diffusion Transformers (DiTs)~\cite{peebles2023scalable}. In parallel, training objectives have evolved from denoising score matching to \emph{flow matching}~\cite{albergo2022building,lipman2023flow,liu2023flow}, which formulates generation as learning a velocity field that transports samples from a simple noise prior to the data distribution, enabling straighter and more efficient generation trajectories.
Building on these, recent models explore different architectural instantiations of DiTs. Stable Diffusion 3 (SD3)~\cite{10.5555/3692070.3692573} introduces the Multimodal Diffusion Transformer (MMDiT), where text and image tokens are processed with separate parameters and interact through joint attention, enabling bidirectional information flow between modalities.
Subsequent models such as FLUX.1 and FLUX.2~\cite{flux2024,flux-2-2025} scale this design and augment MMDiT blocks with single-stream parallel-attention layers operating on concatenated multimodal sequences~\cite{dehghani2023scaling}.
Alternative designs, such as Qwen-Image~\cite{wu2025qwen}, incorporate a frozen multimodal LLM as the text encoder within an MMDiT-like framework, while SANA~\cite{xie2025sana} departs from MMDiT altogether, adopting a single-stream linear-attention DiT paired with a deep-compression autoencoder for high-resolution efficiency. 
In parallel, recent work has focused on reducing sampling cost through distillation. Few-step variants such as FLUX.1-schnell~\cite{flux2024}, FLUX.2-klein~\cite{flux-2-2025}, and SANA-Sprint~\cite{chen2025sana} compress sampling to a handful of steps via trajectory or adversarial distillation. In this work, we study the phenomenon of \emph{massive activations} across this spectrum of modern flow-matching DiTs, spanning MMDiT-based, parallel-attention, and linear-attention architectures, as well as their distilled counterparts.

\tit{Massive Activations in Transformers and DiTs}
\label{sec:related-ma}
Massive activations (MAs) refer to a small subset of hidden-state entries whose values exceed the typical activation scale by several orders of magnitude, often concentrating in fixed channels and acting as implicit bias terms~\cite{sun2024massive}. Originally identified in LLMs and Vision Transformers~\cite{darcet2024vision,sun2024massive}, MAs have recently been observed in DiTs as well, where they concentrate in a handful of fixed channels across image tokens tied to AdaLN modulation~\cite{peebles2023scalable}. Prior work suggests these channels encode limited spatial detail and can degrade dense correspondence unless explicitly controlled~\cite{ganunleashing}. At the same time, MAs are not merely incidental: perturbations to these activations significantly impair \emph{fine-grained detail synthesis} while leaving global semantics largely unchanged, motivating techniques such as Detail Guidance~\cite{gan2026massive}. Complementary analyses of AdaLN conditioning embeddings further reveal a strong form of sparsity, where semantic information is concentrated in a small fraction of dimensions, effectively forming a low-dimensional bottleneck~\cite{phamhidden}.
In contrast to prior work, which primarily studies MAs within individual forward passes or in terms of representation quality, we analyze their role in the generative process of modern MMDiTs. Specifically, we leverage the dual-stream structure of MMDiT architectures to examine image and text streams separately, and we focus on how massive activations behave \emph{across multiple generation trajectories}, enabling a different perspective on their functional role.
\section{Massive Activations as a Sparse Semantic Subspace}
\label{sec:method}

We study the role of massive activations (MAs) in DiTs and analyze their structure and effect on generation. Our analysis proceeds in three steps: (i) we identify their functional importance via channel disruption, (ii) we reveal their spatial organization through clustering, and (iii) we demonstrate their cross-generation role via activation transport. Together, these results provide a unified view of MAs as the main carriers of semantic information.

\tit{Preliminaries}
MMDiTs process two parallel token streams at each layer: an \emph{image stream} \(\mathbf{X}_I^{\ell} \in \mathbb{R}^{N_I \times D}\), consisting of the \(N_I\) latent image tokens being denoised, and an \emph{encoder stream} \(\mathbf{X}_E^{\ell} \in \mathbb{R}^{N_E \times D}\), consisting of the \(N_E\) text tokens that carry the prompt conditioning. The two streams share the channel dimension \(D\) and interact through joint attention within each block. For the image stream, the tokens inherit the spatial layout of the latent image representation. Specifically, if the latent tensor has spatial dimensions \(H_{\mathrm{lat}} \times W_{\mathrm{lat}}\), then,  the number of image tokens is given by \(N_I = H_{\mathrm{lat}} W_{\mathrm{lat}}.\)

We denote by $\mathbf{X} \in \mathbb{R}^{N \times D}$ the hidden activations at a given layer and timestep, where $N$ is the number of total tokens $N_I + N_E$ and $D$ the number of channels. Let \(\alpha \in \{I, E\}\) denote the stream, and let \(N_\alpha\) be the corresponding number of tokens. Given a single sample, we write \(\mathbf{X}_\alpha^{\ell} \in \mathbb{R}^{N_\alpha \times D}\) for the activations of stream \(\alpha\) at layer \(\ell \in \{1,\ldots,L\}\), and \(\mathbf{X}^{\ell}_{\alpha}[n,d]\) entry at token \(n\) and channel \(d\). 

Let us define  \(\{|\mu_d^{\ell}|\}_{d=1}^{D}\) the channel-wise mean distribution at layer $\ell$. This distribution is highly concentrated, and there exists a small subset of channels that accounts for most of the total activation magnitude.
To define this set, we compute a channel-wise importance score, instantiated as the absolute channel mean \(|\mu_{d}^{\ell}|\) at layer \(\ell\). We then select the top-\(k\) and bottom-\(k\) according to this score for each layer independently as:
\begin{equation}
\mathcal{T}_{k}^{\ell}
=
\text{top-}k
\left(\{|\mu_{d}^{\ell}|\}_{d=1}^{D}\right),
\qquad
\mathcal{B}_{k}^{\ell}
=
\text{bottom-}k
\left(\{|\mu_{d}^{\ell}|\}_{d=1}^{D}\right).
\label{eq:top_k_bottom_k}
\end{equation}
We refer to the activations in $\mathcal{T}_k^{\ell}$ as \textit{massive activations}, as they account for most of the total activation magnitude, while the complementary set $\mathcal{B}_k^{\ell}$ contributes negligibly. Further, we define \textbf{ $\mathcal{C}$} as a set of channels corresponding to either $\mathcal{T}_k^{\ell}$ or $\mathcal{B}_k^{\ell}$.
All channel statistics are computed within a single sample, across the token dimension of the given stream layer \(\ell\) in stream \(\alpha\). These are per-sample, per-layer, per-stream, per-channel quantities; no cross-sample averaging is performed.

\subsection{Channel Ablation as an Interventional Probe}
\label{sec:heavy-tailed}
Building on the concentration of channel statistics in Eq.~\ref{eq:top_k_bottom_k}, we investigate their role during generation.
To this end, we extend prior work~\cite{phamhidden} with an interventional probe for modern MMDiT architectures that independently evaluates the role of their two information streams. We call this pipeline \emph{channel disruption}: selectively zeroing a subset of channels to measure their impact on generation.

Formally, given activations $\mathbf{X} \in \mathbb{R}^{N \times D}$, we define the disrupted activations as
$\widetilde{\mathbf{X}} = \mathbf{X} \odot \mathbf{m}$, 
where $\mathbf{m} \in \{0,1\}^D$ is a binary mask that zeros the channels in $\mathcal{C}$. 
The intervention is applied independently at each layer, timestep, and stream.

\begin{figure}[t]
    \centering
    \includegraphics[width=0.98\linewidth]{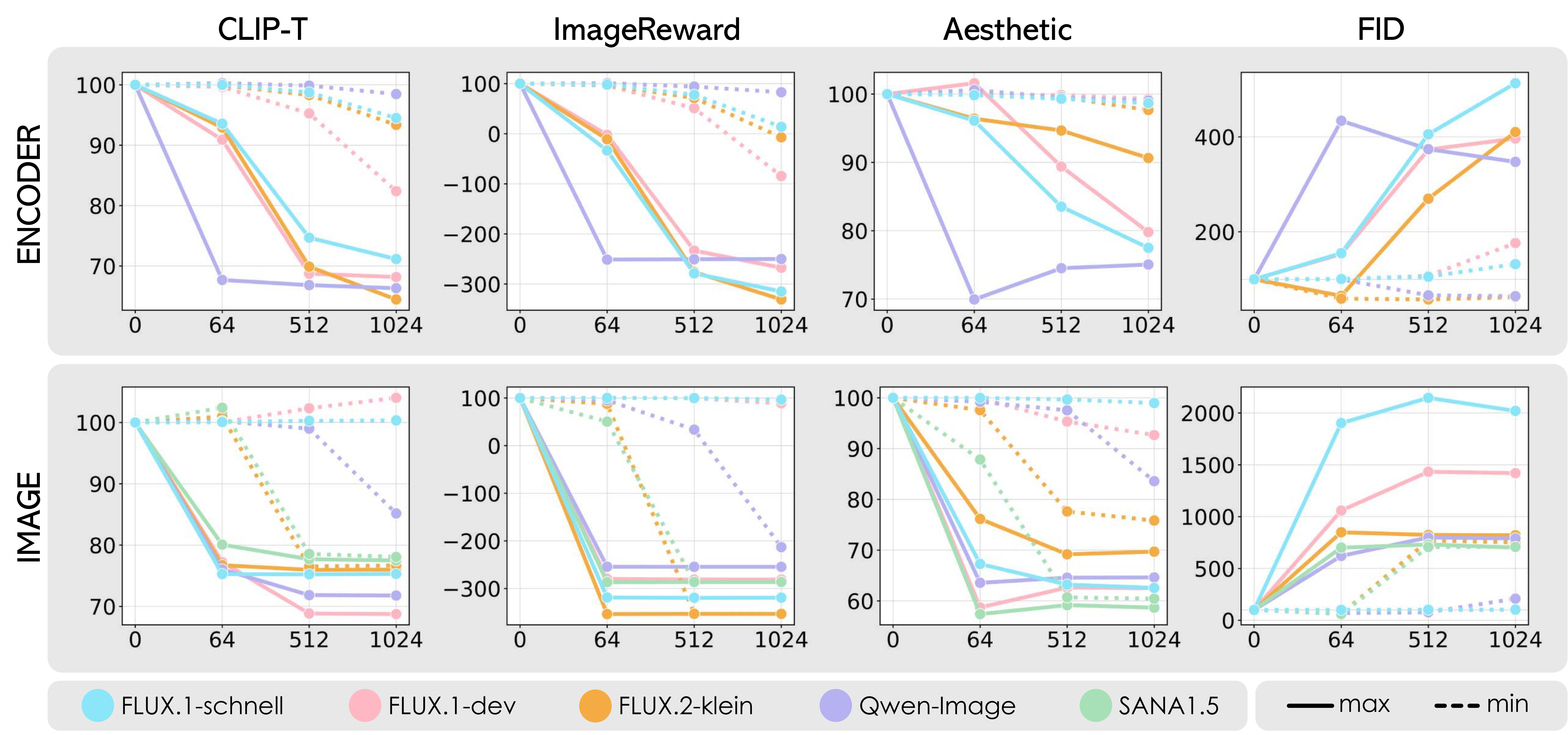}
    \vspace{-0.15cm}
    \caption{Effect of channel disruption across models, metrics, and streams. Each plot reports performance as a function of the number of disrupted channels ($k$), expressed as a percentage relative to the undisrupted baseline (100). Solid lines correspond to disrupting the top-$k$ channels ($\mathcal{T}_k$), while dashed lines correspond to disrupting the bottom-$k$ channels ($\mathcal{B}_k$). Rows separate interventions on the encoder stream (top) and image stream (bottom), while columns report different evaluation metrics.
    }
    \label{fig:disruption}
    \vspace{-0.3cm}
\end{figure}

\tit{Emergent Property}
We evaluate channel disruption across five models, namely FLUX.1-schnell, FLUX.1-dev, FLUX.2-klein, Qwen-Image, and SANA1.5\footnote{Note that SANA1.5 is based on a linear DiT, and therefore uses a single generation stream.}. We apply disruption at all layers and timesteps, and report (Fig.~\ref{fig:disruption}) results on ImageNet-1k~\cite{imagenet15russakovsky} using standard alignment (CLIP-T~\cite{hessel2021clipscore}, ImageReward~\cite{xu2023imagereward}), quality (Aesthetic Score~\cite{schuhmann2022laion}), and distributional (FID~\cite{heusel2017gans}) metrics. Each metric is reported as a percentage of its value on the corresponding un-disrupted generation. We refer the reader to the Appendix for further details and GenAI-Bench~\cite{li2024genai} results.

Two main findings emerge. First, we extend the observation of~\cite{phamhidden}, originally reported on class-conditioned DiTs, to state-of-the-art text-conditioned MMDiT models: disrupting top-ranked channels causes a sharp degradation across models and metrics, whereas removing an equally-sized set of low-ranked channels has only a limited effect. This further supports the view that generative computation is concentrated in a small subset of high-statistic channels.
More importantly, the effect of disruption differs across the two streams. In both streams, suppressing top-ranked channels leads to a comparable drop in CLIP-T and ImageReward, indicating that both streams carry information necessary to produce images aligned with the prompt. However, disrupting the image stream causes a  larger deterioration in FID and Aesthetic score signaling heavier unnatural artifacts. By contrast, disrupting the encoder stream yields a milder degradation, suggesting that it primarily affects prompt alignment while leaving the ability of the models to generate plausible images comparatively intact.

\subsection{Spatial Structure of Image-Stream Channels}
\label{sec:spatial-structure}
While MAs are defined along the channel dimension, we now investigate how they induce structured patterns across the token axis. To expose this structure, we restrict the activations to the top-$k$ selected channels and analyze their spatial organization, to see if salient concepts are highlighted by salient activations.
Specifically, we focus on the image stream, whose tokens inherit the spatial layout of the latent image grid. Let $\mathbf{X}_{I,{\mathcal{C}}} \in \mathbb{R}^{N_I \times k}$ denote the activations restricted to a selected set of channels $\mathcal{C}$, where $\mathcal{C}$ corresponds to either $\mathcal{T}_k^{\ell}$ or $\mathcal{B}_k^{\ell}$ as in Eq. \ref{eq:top_k_bottom_k}.

To extract spatial structure from these activations, we partition tokens based on their feature representations in the restricted subspace. Concretely, we apply $K$-means clustering with $K=2$ to the channels of $\mathbf{X}_{I,\mathcal{C}}$, assigning each token to one of two clusters according to its feature vector in $\mathbb{R}^k$.

\begin{figure}[t]
    \centering
    \includegraphics[width=0.98\linewidth]{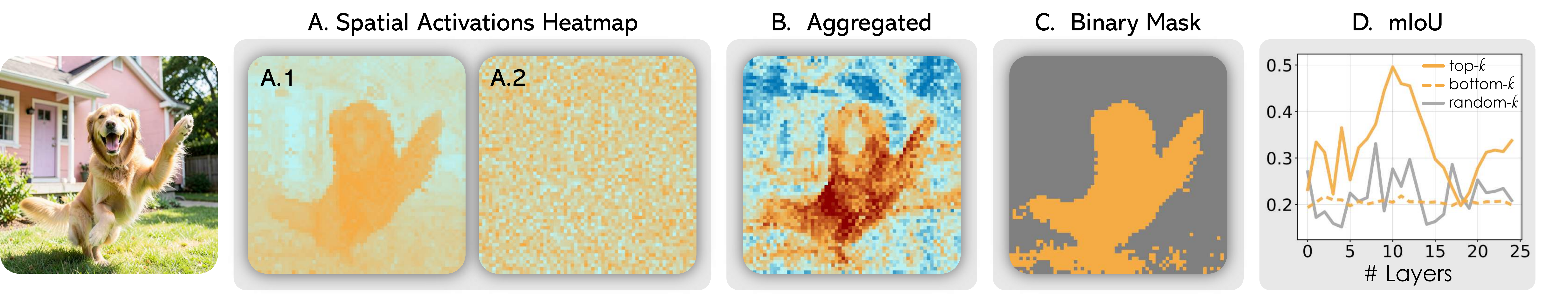}
    \vspace{-0.15cm}
    \caption{Spatial structure of MAs on FLUX.2-klein model. (A) Channel-wise activation maps for high- (A.1) and low-importance (A.2) channels: massive channels exhibit coherent, subject-aligned patterns, while low-importance channels are diffuse. (B) Aggregating activations over top-$k$ channels yields a structured heatmap aligned with salient regions. (C) Clustering these activations produces a binary mask that closely captures the main subject. (D) Quantitative evaluation (mIoU) shows that masks derived from MAs (bold orange) are more aligned with reference segmentation across layers, differently from low-importance activations (dash orange) and random activations (grey).}
    \label{fig:masks}
    \vspace{-0.3cm}
\end{figure}

Each cluster is classified as salient or non-salient based on a score reflecting the strength of the selected activations across its tokens. Specifically, we apply min-max normalization on $\mathbf{X}_{I,{\mathcal{C}}}$ along the channel dimension so that the $k$ spatial maps have comparable scale, denoting the result $\hat{\mathbf{X}}_{I,{\mathcal{C}}}$. Let $\{C_0, C_1\}$ denote the two clusters obtained by $K$-means, where each $C_j \subseteq \{1,\dots,N_I\}$ is the set of tokens assigned to cluster $j$. For each token $n$ we aggregate the normalized activations across the selected channels and compute the per-cluster average as:
\begin{equation}
    \bar{s}_j = \frac{1}{|C_j|}\sum_{n \in C_j} s_n \quad \text{for} \quad j \in \{0,1\}, \quad \text{where} \quad
    s_n = \sum_{c \in \mathcal{C}} \hat{X}_I[n,c].
    \label{eq:aggregation_activations_mask}
\end{equation}
The cluster with the highest average score is assigned to the foreground class, the other to the background. This induces a binary mask $\mathbf{p} \in \{0,1\}^{N_I}$ with $p_n = 1$ if token $n$ belongs to the foreground cluster and $p_n = 0$ otherwise. 
Fig.~\ref{fig:masks}~(B) shows a representative example of the normalized activation heatmap computed via Eq.~\ref{eq:aggregation_activations_mask}, alongside the corresponding binary mask in Fig.~\ref{fig:masks}~(C).

\tit{Emergent Property} 
To evaluate whether the clusters induced by massive activations correspond to semantically meaningful regions, we frame the task as \emph{dichotomous segmentation}. We compute mIoU between the binary mask from the last denoising timestep and a pseudo-ground-truth from BiRefNet~\cite{zheng2024bilateral} on the final generated image, averaged over the 1{,}600 prompts of GenAI-Bench~\cite{li2024genai}.

Fig.~\ref{fig:masks}~(D) reports this metric on FLUX.2-klein for three channel-selection strategies with $k=12$, kept fixed for all subsequent experiments. Top-$k$ consistently dominates bottom-$k$ and random-$k$ at every layer, peaking at mIoU $\approx 0.5$ at layer~10, while bottom-$k$ remains flat around $0.2$ and random-$k$ falls in between, indicating that semantic spatial structure is concentrated in a small, identifiable subset of channels rather than being a generic property of the activations.
Together with the qualitative visualization in Fig.~\ref{fig:masks}~(A--C), these results reveal an emergent property: activations in $\mathcal{T}_k^{\ell}$ concentrate semantic information in space, producing structured maps that align with foreground objects and background regions, whereas $\mathcal{B}_k^{\ell}$ yields diffuse, noise-like patterns. This shows that spatial semantic structure is not uniformly distributed across channels, but is concentrated in a small, identifiable set of massive activations. We refer the reader to the Appendix for further quantitative evaluations.

\subsection{Channel-Selective Activation Transport}
\label{sec:merging}

\begin{figure}[t]
    \centering
    \includegraphics[width=0.98\linewidth]{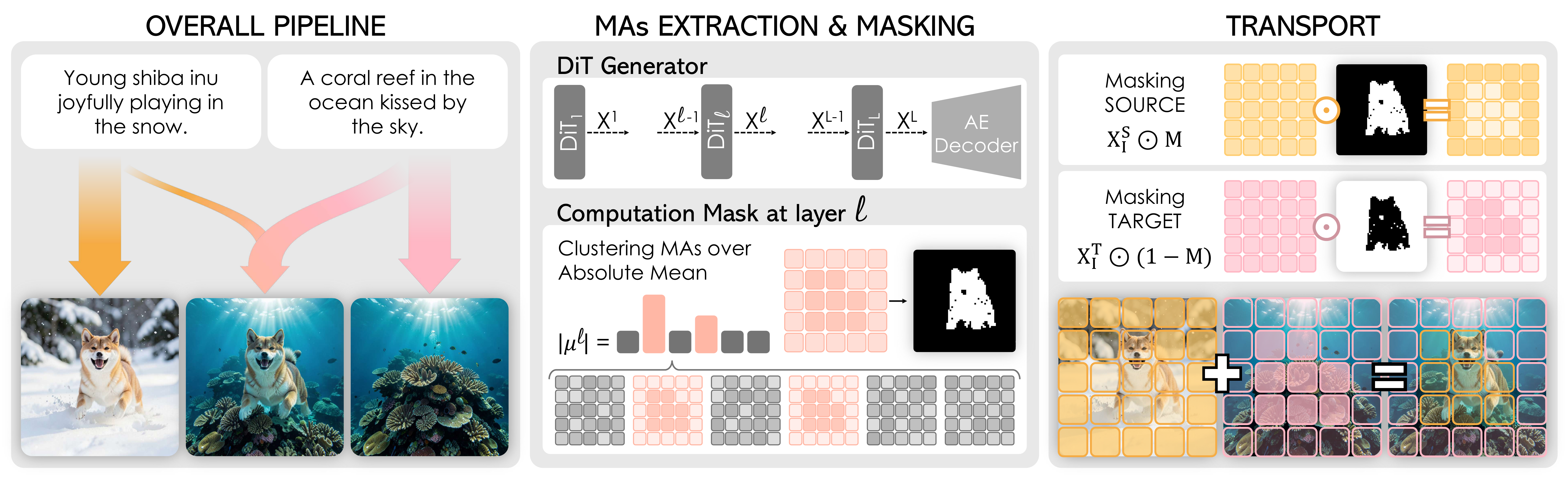}
    \vspace{-0.2cm}
    \caption{Activation transport via MAs. (Left) Overall pipeline: two prompts are used to generate a source and a target image from the same DiT. (Center) Top-$k$ channels are identified in the source trajectory and used to derive a spatial mask. (Right) Source activations are injected into the target generation within the masked region, while the remaining channels are preserved.} 
    \label{fig:method}
    \vspace{-0.3cm}
\end{figure}

Channel disruption shows that a small subset of high-absolute-mean channels has a disproportionate effect on the final generation. We now move from \emph{removing} these channels to \emph{transferring} them across prompts. The goal is to test whether MAs are not only indispensable to the generation process, but also carry prompt-specific semantic information that can be reused in another generation.

We consider two generations obtained from the same initial noise but different prompts, and denote their hidden activations at a given layer by $\mathbf{X}^S$ (source) and $\mathbf{X}^T$ (target), with $\mathbf{X}^S,\mathbf{X}^T \in \mathbb{R}^{N \times D}$. 

\tit{Channel-Selective Replacement}
Instead of zeroing out selected channels, we replace them in the target generation with those from the source. Specifically, using the channel mask $\mathbf{m}\in\{0,1\}^D$ defined in Sec.~\ref{sec:heavy-tailed}, the merged activations are defined as:
\begin{equation}
\widetilde{\mathbf{X}}^T
=
\mathbf{X}^T \odot (1-\mathbf{m})
+
\mathbf{X}^S \odot \mathbf{m},
\label{eq:channel_replace}
\end{equation}
where $\mathbf{m}$ is broadcast across tokens. This channel-wise replacement applies to both~encoder and image streams, injecting sparse source activations into the target while preserving other channels.

\tit{Spatially Selective Replacement in the Image Stream}
The emergent spatial interpretation of activations (Sec.~\ref{sec:spatial-structure}) enables localized transport using the structure induced by massive activations. We reuse the spatial mask $\mathbf{p} \in \{0,1\}^{N_I}$ from Sec.~\ref{sec:spatial-structure}, where $N_I$ is the number of image tokens.

We combine token mask $\mathbf{p}$ with the channel mask $\mathbf{m}$ to obtain a joint channel-spatial mask:
\begin{equation}
\mathbf{M} = \mathbf{p}\mathbf{m}^\top \in \{0,1\}^{N_I \times D}.
\end{equation}

The image-stream activations are then updated as:
\begin{equation}
\widetilde{\mathbf{X}}_I^T
=
\mathbf{X}_I^T \odot (1-\mathbf{M})
+
\mathbf{X}_I^S \odot \mathbf{M},
\label{eq:image_replace}
\end{equation}
while the encoder stream is updated using channel-only replacement (Eq.~\ref{eq:channel_replace}). In the image stream, the joint mask $\mathbf{M}$ replaces the channel mask $\mathbf{m}$, ensuring that replacement occurs only on selected channels within selected spatial regions. We apply this at a selected set of layers at all denoising steps during generation. In practice, intervening at a subset of intermediate layers is already sufficient to induce consistent semantic effects. Fig.~\ref{fig:method} depicts an overview of the transport mechanism.

\begin{figure}[t]
    \centering
    \includegraphics[width=0.98\linewidth]{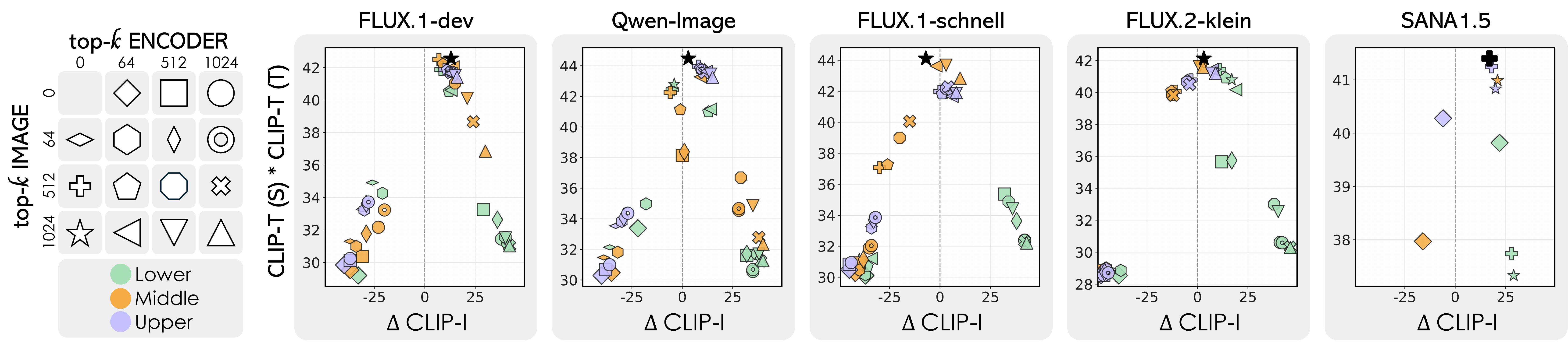}
    \vspace{-0.15cm}
    \caption{Semantic effect of activations transport across models and layer regimes. Each point represents a configuration, where $\Delta$CLIP-I measures the shift toward the source prompt, and CLIP-T(S)*CLIP-T(T) captures joint semantic preservation. Colors indicate the layer regime where the intervention is applied (lower, middle, upper), while marker shapes encode the number of transported channels ($k$) and the stream (encoder or image). Interventions in the middle layers achieve better trade-offs, suggesting that semantic transport is most effective in middle stages of the network.}
    \label{fig:best_conf_prompt_prompt}
    \vspace{-0.4cm}
\end{figure}

\tit{Emergent Property} 
We analyze the effect of activation transport in embedding space. Fig.~\ref{fig:best_conf_prompt_prompt} reports the semantic shift induced by transporting top-$k$ MAs across models, layers, and configurations. Each point corresponds to a transport setup. The horizontal axis, $\Delta\text{CLIP-I} = \text{CLIP-I (S)} - \text{CLIP-I (T)}$, measures the relative similarity of the merged image to the source versus the target, with values near zero indicating merges that integrate both prompts rather than collapsing to one. The vertical axis captures the joint preservation of source and target semantics through CLIP-T. We conduct this analysis on 703 GenAI-Bench~\cite{li2024genai} prompts with a \texttt{scene} attribute to guarantee that a salient subject and a meaningful background are present, ensuring 
coherent merges. We randomly pair each prompt with five others to form 3,515 source-target pairs and report results across the generators used in the previous analyses.
Across all models, three consistent patterns emerge. First, top-$k$ controls the horizontal position: larger top-$k$ transfers more activations from the source and shifts points rightward along $\Delta$CLIP-I, while small top-$k$ keeps generations closer to the target. Second, middle-layer configurations (orange) more often lie in the upper region than lower (green) or upper (purple) layers, suggesting that transport at mid-depths preserves better joint alignment; we therefore focus on mid-layer configurations. Third, the encoder stream contributes little; indeed, configurations with encoder top-$k = 0$ tend to match or outperform those that also transport encoder features.

The configuration achieving $\Delta$CLIP-I $\approx 0$ and the highest CLIP-T(S)*CLIP-T(T) is applied at middle layers with encoder top-$k = 0$ and image top-$k = 1024$ on all models but SANA1.5, where the optimum shifts to image top-$k = 512$. We adopt these settings in subsequent experiments.
These consistent trends suggest activation transport induces structured semantic directions, rather than random perturbations, with MAs operating in a controllable subspace.
\section{Use Cases}
The observations suggest that MAs are a lightweight, architecture-agnostic interface for semantic~control in DiTs. We illustrate two practical use cases: \emph{prompt-to-prompt semantic transport}, where activations are transplanted between 
different, text-conditioned generations and \emph{image-conditioned semantic transport}, where salient subject semantics are extracted and re-injected into new generations.

\label{sec:experiments}
\subsection{Text-Conditioned Semantic Transport}
\label{sec:use_case_p2p}
We consider a setting where two images are generated from the same initial noise but different prompts, denoted $x^S$ and $x^T$. Using the massive activations masks and channel selection described in the previous sections, we transplant a set of MAs from the trajectory of $x^S$ into that of $x^T$, to transfer semantic attributes while preserving the structure of the target image. 

\tit{Experimental Setup}
Following the setup of Sec.~\ref{sec:merging}, we evaluate on the 3,515 prompt pairs from GenAI-Bench across the five DiT-based models. Table~\ref{tab:prompt_prompt} reports CLIP-T, CLIP-I and DINO-I, computed independently against $x^S$ and $x^T$ and combined as a product (S*T) to capture joint fidelity. High scores therefore require simultaneously preserving characteristics from both source and target; accordingly, we focus our comparisons on the combined product rather than the individual metrics.

\tit{Baselines}
We compare our best configuration against three alternatives. First, we consider linear interpolation between the two prompts, applied either to DiT activations at every timestep -- across all layers or restricted to those used in our identified setting -- or to autoencoder latents at each timestep. Second, to assess the role of spatial selection, we include an ablation of our MAs substitution method without the spatial mask (Sec.~\ref{sec:merging}). Finally, as an upper bound that bypasses multi-prompt combination by operating in the natural regime of the generator, we report results obtained by merging the two prompts into a single description using Gemma3-4B~\cite{gemma3_2025} and generating one image from it.

\begin{table}[t]
  \centering
  \setlength{\tabcolsep}{0.45em}
  \caption{Prompt-to-prompt composition results. We compare strategies across models, reporting CLIP-T, CLIP-I, and DINO-I scores for source (S), target (T), and their joint preservation (S*T).}
  \vspace{-0.1cm}
  \resizebox{\linewidth}{!}{%
  \begin{tabular}{ll c c c c c c c c c c c c c c c}
    \toprule
      &  &  &  & &  \multicolumn{3}{c}{\textbf{CLIP-T}} & &  \multicolumn{3}{c}{\textbf{CLIP-I}}  & &  \multicolumn{3}{c}{\textbf{DINO-I}} \\
     \cmidrule{6-8} \cmidrule{10-12} \cmidrule{14-16}
     &  & \textbf{Mask} & \textbf{Layers / Steps}& & \textbf{S*T} & \textbf{S} & \textbf{T} & &  \textbf{S*T} & \textbf{S} & \textbf{T} & &  \textbf{S*T} & \textbf{S} & \textbf{T} \\
    \midrule
    \multirow{6}{*}{\rotatebox{90}{\makecell{FLUX.1-schnell}}}
     & \cellcolor{lightgray}Single Prompt & \cellcolor{lightgray}-  & \cellcolor{lightgray}-& \cellcolor{lightgray} & \cellcolor{lightgray}46.5 & \cellcolor{lightgray}\textcolor{tablegray}{70.3} & \cellcolor{lightgray}\textcolor{tablegray}{66.3} & \cellcolor{lightgray} & \cellcolor{lightgray}52.7 & \cellcolor{lightgray}\textcolor{tablegray}{74.5} & \cellcolor{lightgray}\textcolor{tablegray}{70.8} & \cellcolor{lightgray} &  \cellcolor{lightgray}15.5 & \cellcolor{lightgray}\textcolor{tablegray}{45.4} & \cellcolor{lightgray}\textcolor{tablegray}{34.2} \\
     & Interpolation & \cmark & all & & 31.7 & \textcolor{tablegray}{83.2} & \textcolor{tablegray}{38.1} & &  53.2 & \textcolor{tablegray}{99.5} & \textcolor{tablegray}{53.5} & &  2.7 & \textcolor{tablegray}{99.0} & \textcolor{tablegray}{2.7} \\
     & Latent Interpolation & - & all & & 42.2 & \textcolor{tablegray}{63.7} & \textcolor{tablegray}{66.2} & &  52.2 & \textcolor{tablegray}{71.3} & \textcolor{tablegray}{73.2} & &  18.5 & \textcolor{tablegray}{39.8} & \textcolor{tablegray}{46.5} \\
     & Interpolation & \cmark  & mid & & 37.4 & \textcolor{tablegray}{79.0} & \textcolor{tablegray}{47.3} & &  53.3 & \textcolor{tablegray}{90.7} & \textcolor{tablegray}{58.8} & &  9.3 & \textcolor{tablegray}{80.1} & \textcolor{tablegray}{11.6} \\
     & MAs & - & mid & & 32.4 & \textcolor{tablegray}{83.0} & \textcolor{tablegray}{39.0} & &  53.0 & \textcolor{tablegray}{98.0} & \textcolor{tablegray}{53.4} & &  3.0 & \textcolor{tablegray}{96.3} & \textcolor{tablegray}{3.1} \\
     & \cellcolor{OurColor}\textbf{MAs (Ours)} & \cellcolor{OurColor}\cmark & \cellcolor{OurColor}mid & \cellcolor{OurColor}& \cellcolor{OurColor}\textbf{44.1} & \cellcolor{OurColor}\textcolor{tablegray}{62.8} & \cellcolor{OurColor}\textcolor{tablegray}{70.3} & \cellcolor{OurColor}&  \cellcolor{OurColor}\textbf{55.2} & \cellcolor{OurColor}\textcolor{tablegray}{71.2} & \cellcolor{OurColor}\textcolor{tablegray}{77.6} & \cellcolor{OurColor}&  \cellcolor{OurColor}\textbf{20.1} & \cellcolor{OurColor}\textcolor{tablegray}{42.4} & \cellcolor{OurColor}\textcolor{tablegray}{47.5} \\
     \midrule
    \multirow{6}{*}{\rotatebox{90}{\makecell{FLUX.1-dev}}}
     & \cellcolor{lightgray}Single Prompt & \cellcolor{lightgray}- & \cellcolor{lightgray}- & \cellcolor{lightgray} & \cellcolor{lightgray}44.2 & \cellcolor{lightgray}\textcolor{tablegray}{68.7} & \cellcolor{lightgray}\textcolor{tablegray}{64.4} &\cellcolor{lightgray} &  \cellcolor{lightgray}54.9 & \cellcolor{lightgray}\textcolor{tablegray}{76.0} & \cellcolor{lightgray}\textcolor{tablegray}{72.2} & \cellcolor{lightgray}&  \cellcolor{lightgray}14.4 & \cellcolor{lightgray}\textcolor{tablegray}{43.9} & \cellcolor{lightgray}\textcolor{tablegray}{32.9} \\
     & Interpolation & \cmark & all & & 31.2 & \textcolor{tablegray}{74.0} & \textcolor{tablegray}{42.2} & & 43.4 & \textcolor{tablegray}{79.7} & \textcolor{tablegray}{54.5} & & 2.4 & \textcolor{tablegray}{62.4} & \textcolor{tablegray}{3.8}\\
     & Latent Interpolation & - & all & & 30.6 & \textcolor{tablegray}{80.2} & \textcolor{tablegray}{38.1} & &  57.1 & \textcolor{tablegray}{100} & \textcolor{tablegray}{57.1} & &  4.4 & \textcolor{tablegray}{100} & \textcolor{tablegray}{4.4} \\
     & Interpolation & \cmark & mid & & 32.6 & \textcolor{tablegray}{78.4} & \textcolor{tablegray}{41.6} & &  49.6 & \textcolor{tablegray}{88.4} & \textcolor{tablegray}{56.1} & &  3.2 & \textcolor{tablegray}{76.5} & \textcolor{tablegray}{4.2} \\
     & MAs & - & mid & & 30.9 & \textcolor{tablegray}{80.4} & \textcolor{tablegray}{38.5} & &  57.0 & \textcolor{tablegray}{99.9} & \textcolor{tablegray}{57.0} & &  4.3 & \textcolor{tablegray}{99.9} & \textcolor{tablegray}{4.3} \\
     & \cellcolor{OurColor}\textbf{MAs (Ours)} & \cellcolor{OurColor}\cmark & \cellcolor{OurColor}mid & \cellcolor{OurColor}& \cellcolor{OurColor}\textbf{42.5} & \cellcolor{OurColor}\textcolor{tablegray}{69.2} & \cellcolor{OurColor}\textcolor{tablegray}{61.5} & \cellcolor{OurColor}&  \cellcolor{OurColor}\textbf{58.1} & \cellcolor{OurColor}\textcolor{tablegray}{82.6} & \cellcolor{OurColor}\textcolor{tablegray}{70.4} & \cellcolor{OurColor}&  \cellcolor{OurColor}\textbf{19.2} & \cellcolor{OurColor}\textcolor{tablegray}{61.4} & \cellcolor{OurColor}\textcolor{tablegray}{31.3} \\
     \midrule
    \multirow{6}{*}{\rotatebox{90}{\makecell{FLUX.2-klein}}}
     & \cellcolor{lightgray}Single Prompt & \cellcolor{lightgray}- & \cellcolor{lightgray}- & \cellcolor{lightgray}& \cellcolor{lightgray}43.7 & \cellcolor{lightgray}\textcolor{tablegray}{67.3} & \cellcolor{lightgray}\textcolor{tablegray}{65.0} & \cellcolor{lightgray}&  \cellcolor{lightgray}53.3 & \cellcolor{lightgray}\textcolor{tablegray}{74.0} & \cellcolor{lightgray}\textcolor{tablegray}{72.1} & \cellcolor{lightgray}& \cellcolor{lightgray}16.9 & \cellcolor{lightgray}\textcolor{tablegray}{46.1} & \cellcolor{lightgray}\textcolor{tablegray}{36.7} \\
     & Interpolation & \cmark & all & &  30.2 & \textcolor{tablegray}{81.0} & \textcolor{tablegray}{37.3} & & 52.7 & \textcolor{tablegray}{99.1} & \textcolor{tablegray}{53.2} & & 3.2 & \textcolor{tablegray}{98.7} & \textcolor{tablegray}{3.2} \\
     & Latent Interpolation & - & all & &  31.1 & \textcolor{tablegray}{81.0} & \textcolor{tablegray}{38.4} & & 52.4 & \textcolor{tablegray}{98.0} & \textcolor{tablegray}{53.5} & & 3.6 & \textcolor{tablegray}{95.4} & \textcolor{tablegray}{3.7} \\
     & Interpolation & \cmark & mid & & 41.0 & \textcolor{tablegray}{69.8} & \textcolor{tablegray}{58.8} & &  \textbf{55.2} & \textcolor{tablegray}{81.9} & \textcolor{tablegray}{67.4} & &  16.4 & \textcolor{tablegray}{68.4} & \textcolor{tablegray}{24.0} \\
     & MAs & - & mid & & 30.8 & \textcolor{tablegray}{81.0} & \textcolor{tablegray}{38.0} & &  52.7 & \textcolor{tablegray}{98.7} & \textcolor{tablegray}{53.4} & &  3.3 & \textcolor{tablegray}{98.6} & \textcolor{tablegray}{3.4} \\
     & \cellcolor{OurColor}\textbf{MAs (Ours)} & \cellcolor{OurColor}\cmark & \cellcolor{OurColor}mid & \cellcolor{OurColor}& \cellcolor{OurColor}\textbf{41.7} & \cellcolor{OurColor}\textcolor{tablegray}{64.6} & \cellcolor{OurColor}\textcolor{tablegray}{64.5} & \cellcolor{OurColor}& \cellcolor{OurColor}\textbf{55.2} & \cellcolor{OurColor}\textcolor{tablegray}{75.2} & \cellcolor{OurColor}\textcolor{tablegray}{73.4} & \cellcolor{OurColor}& \cellcolor{OurColor}\textbf{19.1} & \cellcolor{OurColor}\textcolor{tablegray}{58.0} & \cellcolor{OurColor}\textcolor{tablegray}{32.9}\\
     \midrule
    \multirow{6}{*}{\rotatebox{90}{\makecell{Qwen-Image}}}
     & \cellcolor{lightgray}Single Prompt & \cellcolor{lightgray}- & \cellcolor{lightgray}- & \cellcolor{lightgray}& \cellcolor{lightgray}\cellcolor{lightgray}44.0 & \cellcolor{lightgray}\textcolor{tablegray}{68.2} & \cellcolor{lightgray}\textcolor{tablegray}{64.5} & \cellcolor{lightgray}&  \cellcolor{lightgray}51.8 & \cellcolor{lightgray}\textcolor{tablegray}{73.7} & \cellcolor{lightgray}\textcolor{tablegray}{70.3} & \cellcolor{lightgray}&  \cellcolor{lightgray}12.5 & \cellcolor{lightgray}\textcolor{tablegray}{40.5} & \cellcolor{lightgray}\textcolor{tablegray}{30.9} \\
     & Interpolation & \cmark & all & & 23.2 & \textcolor{tablegray}{49.5} & \textcolor{tablegray}{46.9} & &  27.9 & \textcolor{tablegray}{53.7} & \textcolor{tablegray}{52.1} & &  0.0 & \textcolor{tablegray}{2.7} & \textcolor{tablegray}{1.3} \\
     & Latent Interpolation & - & all & & 31.2 & \textcolor{tablegray}{82.8} & \textcolor{tablegray}{37.7} & &  \textbf{55.8} & \textcolor{tablegray}{100} & \textcolor{tablegray}{55.8} & &  3.7 & \textcolor{tablegray}{99.9} & \textcolor{tablegray}{3.7} \\
     & Interpolation & \cmark & mid & & 22.8 & \textcolor{tablegray}{48.3} & \textcolor{tablegray}{46.8} & &  26.2 & \textcolor{tablegray}{51.9} & \textcolor{tablegray}{50.5} & &  0.0 & \textcolor{tablegray}{2.0} & \textcolor{tablegray}{1.3} \\
     & MAs & - & mid & & 32.4 & \textcolor{tablegray}{82.8} & \textcolor{tablegray}{39.2} & &  55.2 & \textcolor{tablegray}{98.2} & \textcolor{tablegray}{56.2} & &  4.0 & \textcolor{tablegray}{96.0} & \textcolor{tablegray}{4.2} \\
     & \cellcolor{OurColor}\textbf{MAs (Ours)} & \cellcolor{OurColor}\cmark &\cellcolor{OurColor}mid & \cellcolor{OurColor}& \cellcolor{OurColor}\textbf{44.5} & \cellcolor{OurColor}\textcolor{tablegray}{67.0} & \cellcolor{OurColor}\textcolor{tablegray}{66.4} & \cellcolor{OurColor}&  \cellcolor{OurColor}55.4 & \cellcolor{OurColor}\textcolor{tablegray}{76.0} & \cellcolor{OurColor}\textcolor{tablegray}{72.9} & \cellcolor{OurColor}&  \cellcolor{OurColor}\textbf{18.0} & \cellcolor{OurColor}\textcolor{tablegray}{48.6} & \cellcolor{OurColor}\textcolor{tablegray}{37.1} \\
     \midrule
    \multirow{6}{*}{\rotatebox{90}{\makecell{SANA1.5}}}
     & \cellcolor{lightgray}Single Prompt & \cellcolor{lightgray}- & \cellcolor{lightgray}- & \cellcolor{lightgray}& \cellcolor{lightgray}\cellcolor{lightgray}46.8 & \cellcolor{lightgray}\textcolor{tablegray}{70.0} & \cellcolor{lightgray}\textcolor{tablegray}{66.8} &\cellcolor{lightgray} &  \cellcolor{lightgray}55.6 & \cellcolor{lightgray}\textcolor{tablegray}{76.0} & \cellcolor{lightgray}\textcolor{tablegray}{73.2} & \cellcolor{lightgray}&  \cellcolor{lightgray}18.0 & \cellcolor{lightgray}\textcolor{tablegray}{46.6} & \cellcolor{lightgray}\textcolor{tablegray}{38.6} \\
     & Interpolation & \cmark & all & & 30.2 & \textcolor{tablegray}{63.8} & \textcolor{tablegray}{47.3} & &  34.1 & \textcolor{tablegray}{63.7} & \textcolor{tablegray}{53.6} & &  1.9 & \textcolor{tablegray}{32.6} & \textcolor{tablegray}{5.8} \\
     & Latent Interpolation & - & all & & 32.5 & \textcolor{tablegray}{83.4} & \textcolor{tablegray}{39.0} & &  56.5 & \textcolor{tablegray}{100.0} & \textcolor{tablegray}{56.5} & &  5.4 & \textcolor{tablegray}{99.9} & \textcolor{tablegray}{5.4} \\
     & Interpolation & \cmark & mid & & 32.7 & \textcolor{tablegray}{59.2} & \textcolor{tablegray}{55.3} & &  35.5 & \textcolor{tablegray}{61.4} & \textcolor{tablegray}{57.9} & &  4.00 & \textcolor{tablegray}{23.6} & \textcolor{tablegray}{17.0} \\
     & MAs & - & mid & & 33.0 & \textcolor{tablegray}{83.0} & \textcolor{tablegray}{39.8} & &  56.1 & \textcolor{tablegray}{98.3} & \textcolor{tablegray}{57.1} & &  5.30 & \textcolor{tablegray}{96.9} & \textcolor{tablegray}{5.50} \\
     & \cellcolor{OurColor}\textbf{MAs (Ours)} & \cellcolor{OurColor}\cmark & \cellcolor{OurColor}mid & \cellcolor{OurColor}& \cellcolor{OurColor}\textbf{41.4} & \cellcolor{OurColor}\textcolor{tablegray}{72.8} & \cellcolor{OurColor}\textcolor{tablegray}{56.8} & \cellcolor{OurColor}&  \cellcolor{OurColor}\textbf{55.7} & \cellcolor{OurColor}\textcolor{tablegray}{83.7} & \cellcolor{OurColor}\textcolor{tablegray}{66.6} & \cellcolor{OurColor}&  \cellcolor{OurColor}\textbf{17.0} & \cellcolor{OurColor}\textcolor{tablegray}{65.0} & \cellcolor{OurColor}\textcolor{tablegray}{26.1} \\
    \bottomrule
  \end{tabular}%
  }
  \label{tab:prompt_prompt}
  \vspace{-0.4cm}
\end{table}

\tit{Experimental Results}
Across all backbones, MAs-based transport with spatial masking consistently outperforms interpolation baselines in terms of combined image fidelity.
For instance, on FLUX.1-schnell MAs transport reaches CLIP-I $55.2$ and DINO-I $20.1$ (vs.\ $53.2$ / $2.7$ for vanilla interpolation). Similar margins hold on FLUX.1-dev (CLIP-I $58.1$, DINO-I $19.2$) and FLUX.2-klein (CLIP-I $55.2$, DINO-I $19.1$). Notably, interpolation methods often collapse toward one of the two sources (despite an interpolation $\alpha=0.5$), as reflected by near-zero DINO-I (S*T) scores, whereas our approach preserves meaningful information from both.
The single-prompt upper bound achieves the best text alignment, as expected, since the LLM can fully orchestrate both prompts semantically, leaving the generator with the simpler task of single-prompt generation. However, it consistently underperforms on image-based metrics, as reducing the pair to one prompt sacrifices the identities of $x^S$ and $x^T$. 

In contrast, our identified setting remains within ${\sim}2$ CLIP-T points of this bound, while substantially improving image fidelity (\eg, $+2.5$ CLIP-I and $+4.6$ DINO-I on FLUX.1-schnell), with similar gains across other models. 
Consistent with our earlier observations, spatial masking is crucial for properly localizing the intervention. Removing it causes MA transplantation to propagate beyond the target region, resulting in diminished alignment and image fidelity.
On FLUX.1-schnell, adding the mask increase DINO-I from $3.0$ to $20.1$ and CLIP-T from $32.4$ to $44.1$. This highlights the importance of spatially restricting the intervention to preserve coherent structure.
Overall, these results are consistent with the view that MAs define a prompt-conditioned semantic subspace. Transplanting them steers the generative trajectory along meaningful semantic directions, enabling targeted attribute transfer rather than uncontrolled mixing. Fig.~\ref{fig:prompt_prompt} (left) shows a qualitative example of prompt-to-prompt semantic transport on FLUX.1-dev. More baselines and ablations are shown in the Appendix.

\subsection{Image-Conditioned Semantic Transport}
\label{sec:use_case_i2p}
We extend our analysis to image-conditioned generation, examining whether MAs capture subject-specific semantics not only in text-driven trajectories, but also when conditioning on real images. Unlike the previous use case, where the source trajectory $x^S$ is generated from a text prompt, here it is derived from a reference subject image, while the transport mechanism introduced in Sec.~\ref{sec:merging} remains unchanged: top-$k$ MAs are extracted and injected into the target trajectory $x^T$.
To obtain $x^S$, we rely on the reconstruction technique proposed by TokenVerse~\cite{garibi2025tokenverse}, which learns 
per-example MLPs to predict latent prompt adjustments that steer the generation toward the reference image.
MAs are extracted along this reconstruction trajectory and transported into the target-prompt generation. For FLUX.2-klein, which natively supports image conditioning, we additionally analyze MAs obtained by pairing the subject image with an empty prompt. This allows us to compare representations derived from reconstruction-based inversion with those obtained through direct image conditioning.

\begin{table}[t]
  \centering
  \setlength{\tabcolsep}{0.4em}
  \caption{Image-conditioned transport results. We compare strategies across models, reporting CLIP-T, CLIP-I, and CLIP-I-personalized scores and their joint composition.}
  \vspace{-0.1cm}
  \resizebox{0.92\linewidth}{!}{%
  \begin{tabular}{l l c c c c c}
    \toprule
    \textbf{Generator} & \textbf{Features} & \textbf{CLIP-T*CLIP-I} & \textbf{CLIP-T*CLIP-I$_{\text{pers}}\!$} & \textbf{CLIP-I$_{\text{pers}}$} & \textbf{CLIP-I} & \textbf{CLIP-T} \\
    \midrule
    \rowcolor{lightgray} Qwen-Image-Edit & - & 65.5 & - & \textcolor{tablegray}{-} & \textcolor{tablegray}{74.8} & \textcolor{tablegray}{87.5} \\
    \rowcolor{lightgray} FLUX-Kontext  & - & 66.0 & - & \textcolor{tablegray}{-} & \textcolor{tablegray}{79.3} & \textcolor{tablegray}{83.1} \\
    \rowcolor{lightgray} FLUX.2-klein  & - & 66.8 & - & \textcolor{tablegray}{-} & \textcolor{tablegray}{75.8} & \textcolor{tablegray}{88.2} \\
    \midrule
    \multirow{2}{*}{Qwen-Image}
     & TokenVerse & 58.2 & 58.2 & \textcolor{tablegray}{68.1} & \textcolor{tablegray}{68.0} & \textcolor{tablegray}{85.5} \\
     & \cellcolor{OurColor}\textbf{MAs (Ours)} & \cellcolor{OurColor}\textbf{62.6} & \cellcolor{OurColor}\textbf{65.9} & \cellcolor{OurColor}\textcolor{tablegray}{80.5} & \cellcolor{OurColor}\textcolor{tablegray}{76.5} & \cellcolor{OurColor}\textcolor{tablegray}{81.9} \\
    \midrule
    \multirow{2}{*}{FLUX.1-dev}
     & TokenVerse & 55.3 & 55.3 & \textcolor{tablegray}{67.0} & \textcolor{tablegray}{67.0} & \textcolor{tablegray}{82.5} \\
     & \cellcolor{OurColor}\textbf{MAs (Ours)} & \cellcolor{OurColor}\textbf{58.1} & \cellcolor{OurColor}\textbf{64.7} & \cellcolor{OurColor}\textcolor{tablegray}{85.9} & \cellcolor{OurColor}\textcolor{tablegray}{77.2} & \cellcolor{OurColor}\textcolor{tablegray}{75.3} \\
    \midrule
    \multirow{3}{*}{FLUX.2-klein} & TokenVerse & 63.4 & 62.2 & \textcolor{tablegray}{71.6} & \textcolor{tablegray}{73.0} & \textcolor{tablegray}{86.9} \\
    {FLUX.2-klein} & MAs (TV) & 65.7 & 65.3 & \textcolor{tablegray}{77.1} & \textcolor{tablegray}{77.5} & \textcolor{tablegray}{84.7} \\
     & \cellcolor{OurColor}\textbf{MAs (EDIT)} & \cellcolor{OurColor}\textbf{66.8} & \cellcolor{OurColor}\textbf{66.8} & \cellcolor{OurColor}\textcolor{tablegray}{79.6} & \cellcolor{OurColor}\textcolor{tablegray}{79.6} & \cellcolor{OurColor}\textcolor{tablegray}{83.9} \\
    \midrule
    \multirow{2}{*}{FLUX.1-schnell}
     & TokenVerse & 57.5 & 58.1 & \textcolor{tablegray}{67.6} & \textcolor{tablegray}{67.0} & \textcolor{tablegray}{85.9} \\
     & \cellcolor{OurColor}\textbf{MAs (Ours)} & \cellcolor{OurColor}\textbf{60.8} & \cellcolor{OurColor}\textbf{62.2} & \cellcolor{OurColor}\textcolor{tablegray}{75.5} & \cellcolor{OurColor}\textcolor{tablegray}{73.8} & \cellcolor{OurColor}\textcolor{tablegray}{82.4} \\
    \midrule
    \multirow{2}{*}{SANA1.5}
     & TokenVerse & 56.7 & 55.6 & \textcolor{tablegray}{64.3} & \textcolor{tablegray}{65.4} & \textcolor{tablegray}{86.6} \\
     & \cellcolor{OurColor}\textbf{MAs (Ours)} & \cellcolor{OurColor}\textbf{56.8} & \cellcolor{OurColor}\textbf{63.3} & \cellcolor{OurColor}\textcolor{tablegray}{80.0} & \cellcolor{OurColor}\textcolor{tablegray}{71.9} & \cellcolor{OurColor}\textcolor{tablegray}{79.1} \\
    \bottomrule
  \end{tabular}%
  }
  \label{tab:prompt_img}
  \vspace{-0.1cm}
\end{table}

\begin{figure}[t]
    \centering
    \includegraphics[width=0.98\linewidth]{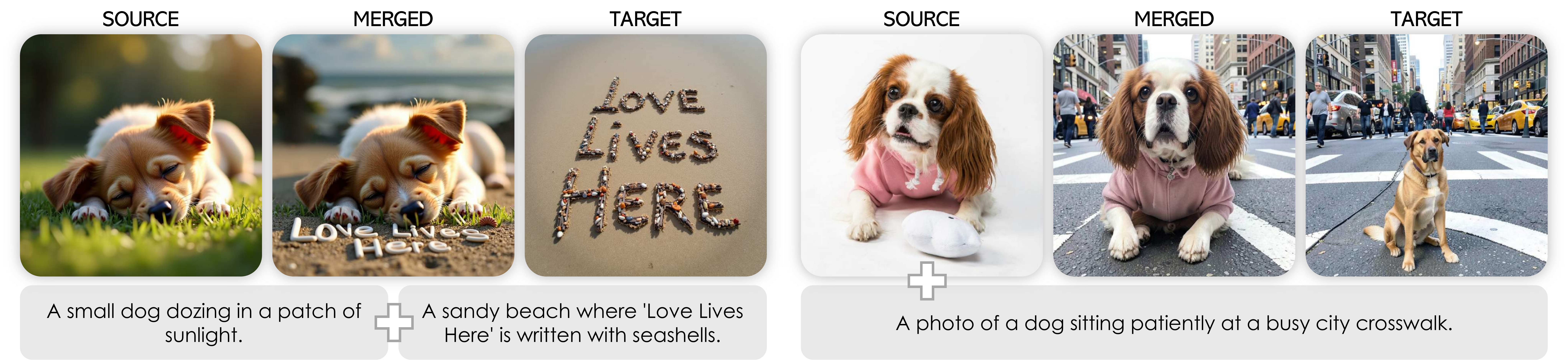}
    \vspace{-0.15cm}
    \caption{Qualitative examples prompt-to-prompt (left) and image-conditioned (right) semantic transport. 
    We refer the user to the Appendix for more qualitative results on all models for both tasks.}
    \label{fig:prompt_prompt}
    \vspace{-0.4cm}
\end{figure}

\tit{Experimental Setup} We evaluate MA-based transport in the image-conditioned setting on DreamBench++~\cite{peng2024dreambench}, a standard personalization benchmark consisting of 150 real subject images paired with prompts in diverse contexts. 
To enable reconstruction, each subject image is first captioned using Qwen-2.5-VL-3B~\cite{qwen2.5}, providing the initial prompt representation on which TokenVerse learns its latent adjustments. Personalization methods typically involve a trade-off between subject fidelity and prompt alignment. To capture this, we report CLIP-I and CLIP-T as measures of image fidelity and text alignment, respectively. Since both TokenVerse and our MA-based approach rely on features extracted from a reconstructed trajectory, we further report CLIP-I$_{\text{pers}}$, computed against the reconstructed subject, to disentangle injection fidelity from reconstruction quality. Finally, we summarize the trade-off using joint metrics CLIP-T*CLIP-I and CLIP-T*CLIP-I$_{\text{pers}}$.

\tit{Baselines} We compare our approach against the native transfer mechanism of TokenVerse, which propagates subject-token activations through the encoder stream across layers to guide the target generation. Moreover, we include dedicated image-editing models designed for personalization as reference upper bounds, namely FLUX.1-kontext~\cite{labs2025flux}, Qwen-Image-Edit~\cite{wu2025qwen} and Flux.2-klein~\cite{flux-2-2025}.

\tit{Experimental Results}
As shown in Table~\ref{tab:prompt_img}, MAs-based feature injection consistently outperforms TokenVerse on the joint personalization metric
CLIP-T*CLIP-I$_{\text{pers}}$.  This improvement is primarily driven by significantly stronger subject preservation: CLIP-I$_{\text{pers}}$ improves by $+12.4$ on
Qwen-Image, $+18.9$ on FLUX.1-dev, $+7.9$ on FLUX.1-schnell, and $+15.7$ on SANA1.5. Importantly, CLIP-T remains in a comparable high range ($75$-$82$), indicating that improved subject fidelity does not come at the expense of prompt alignment. 
On FLUX.2-klein, the two extraction modes (\textit{edit}
and \textit{TV}) yield comparable scores, indicating that MAs-based transport is compatible with both direct image conditioning and reconstruction. Overall, this use case reaches scores comparable to dedicated editing backbones, matching FLUX.2-klein on CLIP-T*CLIP-I ($66.8$) and approaching Qwen-Image-Edit ($62.6$ vs.\ $65.5$), while operating on top of unmodified text-to-image generators and confirming that the properties of MAs established in earlier sections transfer to a downstream personalization setting. Fig.~\ref{fig:prompt_prompt} (right) shows a qualitative example of image-conditioned semantic transport on FLUX.2-klein.
\section{Conclusion}
\label{sec:conclusion}
We have shown that a small subset of channels, called \textit{massive activations}, plays a central role in diffusion transformer generation, exhibiting structured spatial organization and supporting controllable semantic transfer across prompts. These results suggest that massive activations define a sparse subspace through which semantic information is represented and manipulated during generation, reconciling prior views of them as outliers or detail drivers.
Broadly, these findings open questions about the emergence and exploitation of such low-dimensional control structures in generative models.

\clearpage
\section*{Acknowledgments}
We acknowledge CINECA for the availability of high-performance computing resources under the ISCRA initiative, and for funding Evelyn Turri's PhD. This work has been supported by the EU Horizon projects ``ELIAS'' (GA No. 101120237) and ``ELLIOT'' (GA No. 101214398), and by the EuroHPC JU project  ``MINERVA'' (GA No. 101182737).

\bibliographystyle{plain}
\bibliography{bibliography}


\newpage
\appendix

\section{Additional Implementation Details}
\tit{Model Setup}
For the disruption experiment (Sec.~\ref{sec:heavy-tailed}), we generate $5$ images per ImageNet class, following \cite{phamhidden}. This results in a total of 5000 images for each model. 
In text-based experiments, resolution is set to 1024x1024 for all models except Qwen-Image, where it is set to 512x512 for computational constraints, while in image-based ones (Sec.~\ref{sec:use_case_i2p}) also FLUX.1-dev and FLUX.1-schnell are set up at 512x512 resolution. As for the experiments of spatial structures (Sec.~\ref{sec:spatial-structure}), metrics are obtained extracting activations at the last step of generation.

Depending on the depth of each model, we define the Lower, Middle, and Upper layer groups as blocks 0--18, 19--37, 38--56 for FLUX.1-schnell and FLUX.1-dev; 0--7, 8--15, 16--24 for FLUX.2-klein; and 0--19, 20--39, 40--59 for Qwen-Image and SANA1.5, partitioning layers in equal parts. 

To provide a more in-depth analysis, we portray both few-step models and classical multi-step ones, setting FLUX.1-schnell to 1 step, FLUX.2-klein to 4 steps, SANA1.5 to 20 steps, FLUX.1-dev to 28 steps, and Qwen-Image to 28 steps.

\tit{Embedding-based Metrics} For CLIP-based similarity metrics, we employ CLIP~ViT-B/32, and CLIP-T metrics are scaled with a $w=2.5$, as in~\cite{hessel2021clipscore}. For DINO-based similarity metrics, the model used is DINOv2-Base.

\tit{TokenVerse Reconstruction Setup} We optimize each concept for $1400$ steps with learning rate $10^{-4}$, a cosine schedule preceded by $50$ linear warmup steps, zero weight decay, and gradient clipping at norm $1.0$. The training objective is the diffusion loss, with timesteps drawn from a logit-normal distribution. For $50\%$ of the steps, we additionally apply the concept isolation loss of~\cite{garibi2025tokenverse}.

\section{Additional Analysis}
\subsection{Additional Disruption Analysis}
\begin{figure}[b]
\vspace{-0.3cm}
    \centering
    \includegraphics[width=1\linewidth]{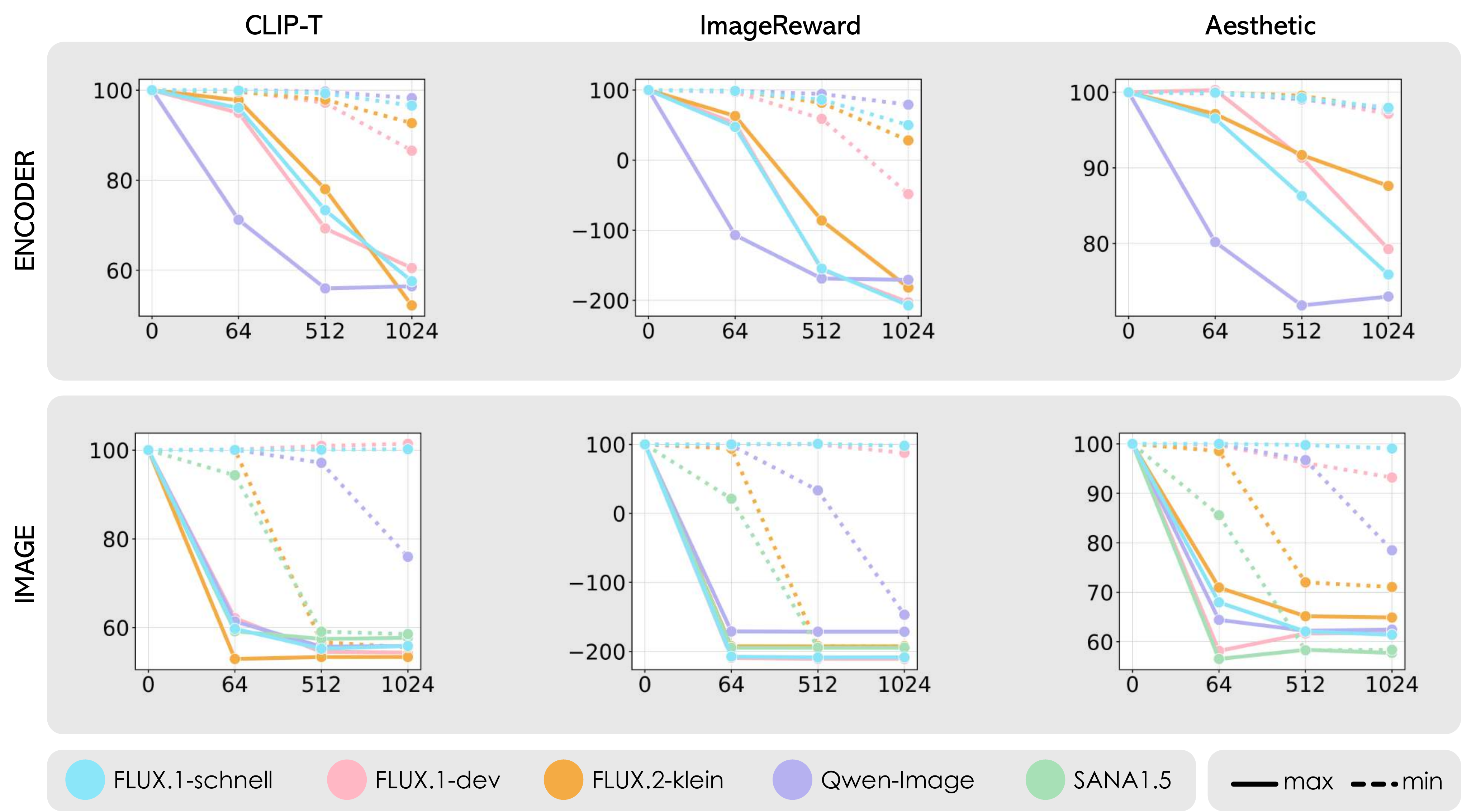}
    \vspace{-0.15cm}
    \caption{Effects of channel disruption on GenAI Bench, with top-$k$ (solid) and bottom-$k$ (dashed) channel selection, with $k$ on the horizontal axis. Results are reported relative the un-disrupted baseline.}
    \label{fig:disruption_genai}
\end{figure}

We provide an additional insight into channel disruption by performing it on images generated on prompts from GenAI-Bench. Fig.~\ref{fig:disruption_genai} shows the obtained results, confirming the observations from Sec~\ref{sec:heavy-tailed}. The same considerations about different behaviors in the image stream and encoder stream also hold when considering complex prompts, that go beyond simple class-like conditioning of ImageNet.

\subsection{Additional Spatial Analysis}
\begin{figure}[t]
    \centering
    \includegraphics[width=0.98\linewidth]{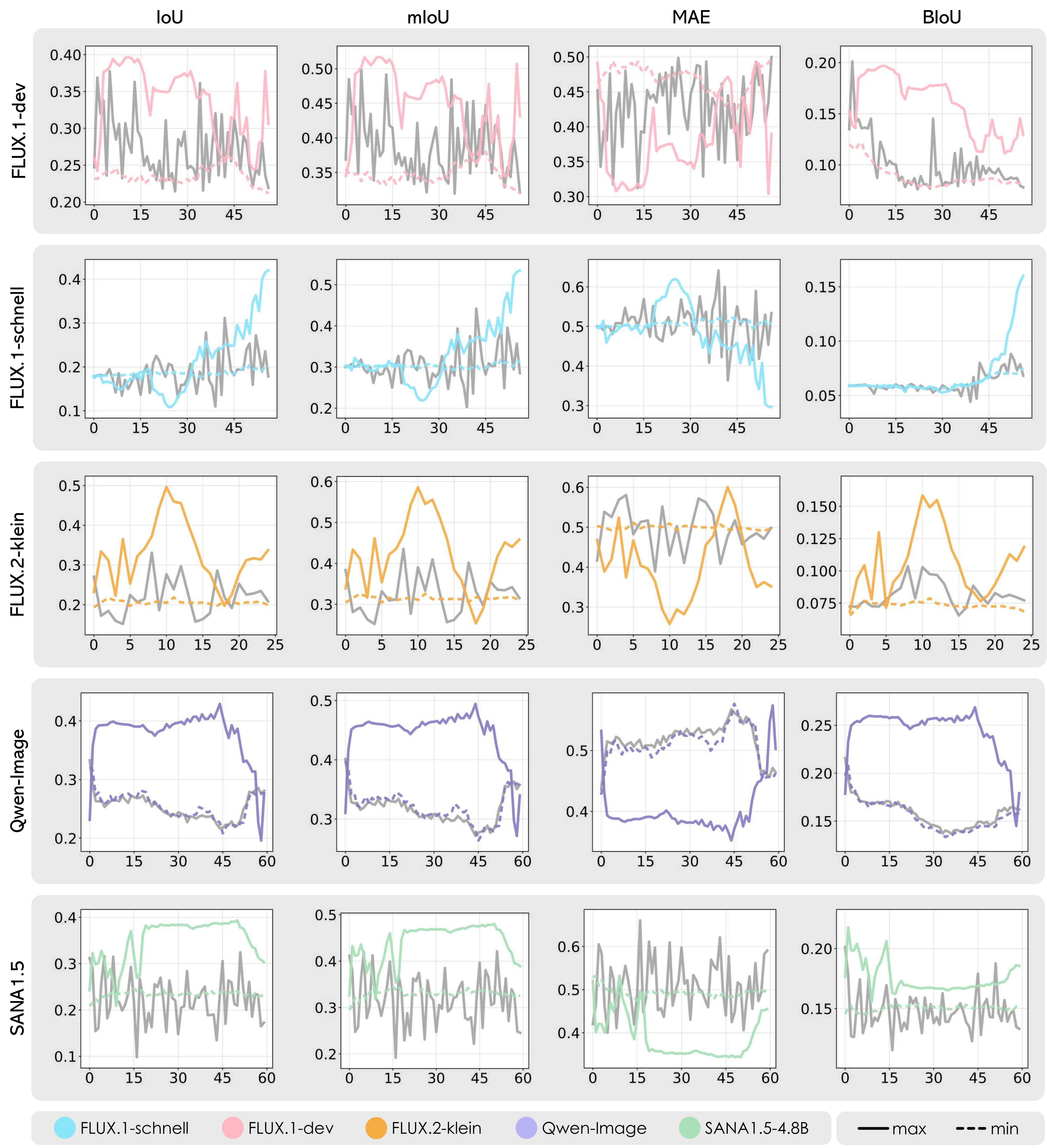}
    \vspace{-0.1cm}
    \caption{Metrics for dichotomous image segmentations across layers for various generative models. Solid lines refer to the use of top-$k$ activations, grey lines to random-$k$ activations, and dotted lines refer to bottom-$k$ activations. Layer indexes are reported in $[0,L-1]$ range.}
    \label{fig:supp_metrics_mask}
    \vspace{-0.3cm}
\end{figure}

Fig.~\ref{fig:supp_metrics_mask} shows an extended analysis of MAs-induced segmentation (as in Sec.~\ref{sec:spatial-structure}) for the five selected models. In particular, we show metrics for dichotomous image segmentation, with pseudo ground truth clusters obtained through~\cite{zheng2024bilateral}. We report standard segmentation metrics: IoU, the intersection over union on the foreground class; mIoU, the mean IoU averaged over foreground and background; MAE, the pixel-level Mean Absolute Error; and bIoU, the IoU restricted to a narrow band around class boundaries.

\begin{figure}[t]
    \centering    \includegraphics[width=0.98\linewidth]{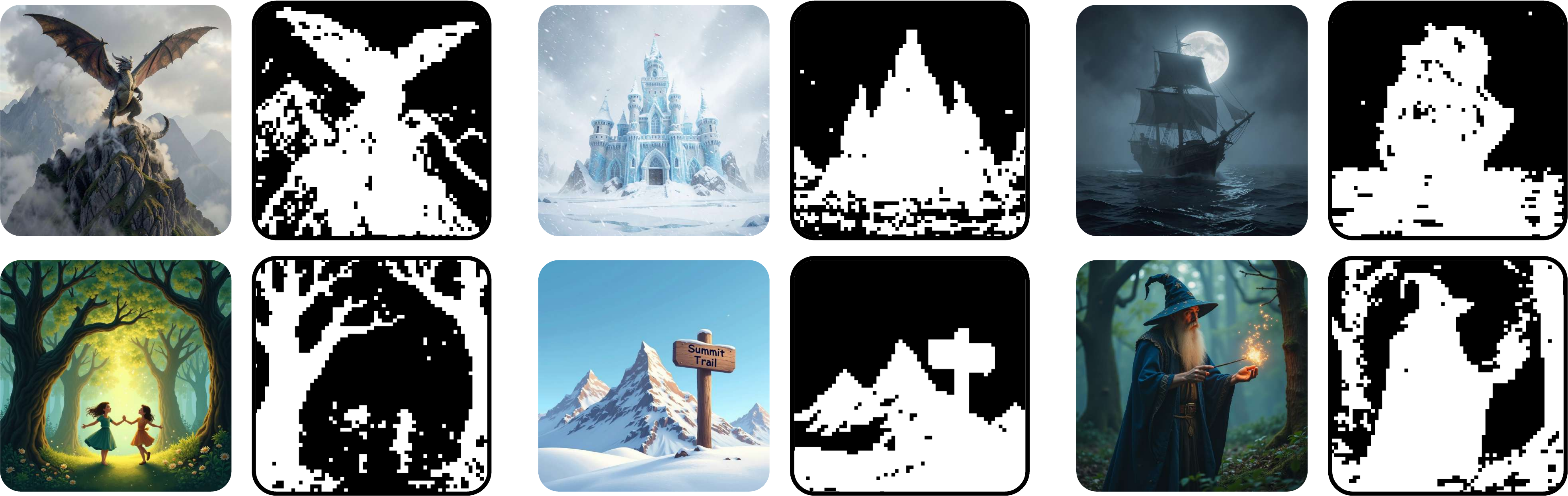}
    \vspace{-0.2cm}
    \caption{Qualitative examples for MAs-based masking extraction, on FLUX.1-dev (top row) and FLUX.2-klein (bottom row)}
    \label{fig:supp_mask_quali}
    \vspace{-0.3cm}
\end{figure}

The pattern is regular: best scores are consistently achieved by top-$k$ activations, while we observe high variability with respect to layer selection. In particular, the best results are achieved at layer~$11$ for FLUX.1-dev, layer~$44$ for Qwen-Image, layer~$10$ for FLUX.2-klein, and layer~$56$ for FLUX.1-schnell. This can be attributed to the architectural and training differences among the considered backbones, which differ in depth and in the way semantic information is consolidated along the network, so that the layer at which subject-level attributes emerge most prominently is model-specific. In contrast, the top-$k$ activation selection consistently identifies the relevant units regardless of where this consolidation occurs.
Finally, Fig.~\ref{fig:supp_mask_quali} shows qualitative examples of MAs-extracted masks, as described in Sec.~\ref{sec:spatial-structure}.

\subsection{Additional Transport Analysis}
\tit{Baselines} We compare MAs against additional interpolation baselines that operate at different stages of the generation pipeline. Interpolation ($\alpha{=}0.5$) is applied to DiT activations either across all transformer layers (\textit{all}) or restricted to the middle layers, as identified as best in Sec.~\ref{sec:merging} as (\textit{mid}). 

Latent Interpolation performs the analogous ($\alpha{=}0.5$) mixing on the autoencoder latents, applied at either the first (\textit{1st}) or last (\textit{last}) denoising step. MAs (Ours) follows the masked, mid layer-step setup described in Sec.~\ref{sec:merging}.

Interpolation collapses toward the source on all five backbones, with CLIP-T (S$*$T) between $22.1$ and $33.9$ and DINO-I (S$*$T) never above $3.3$. Latent interpolation behaves differently: one denoising window (\textit{1st}) yields a balanced regime competitive with MAs, while the other (\textit{last}) collapses to a near-copy of source or target, evidenced by S or T above 80\% paired with the complementary score under 40\%. MAs is the only method that consistently balances source and target alignment.

\tit{Different Feature Selection Criteria} Following the same experimental protocol of Sec.~\ref{sec:merging}, and starting from our best configuration (mid layers, 1024 shifted activations), we report in Tab.~\ref{tab:sup_prompt_prompt_top_k} an extended ablation that jointly varies the activation selection criterion (random, bottom, top) and the mask criterion (random, min, max) across five backbones (FLUX.1-schnell, FLUX.1-dev, FLUX.2-klein, Qwen-Image, SANA1.5) and three similarity metrics (CLIP-T, CLIP-I, DINO). For each metric we report the joint score S*T alongside the source-only (S) and target-only (T) components, the latter shown in gray as they are reported for completeness as they are not relevant on their own. The pattern is consistent across all backbones and metrics: Top activation selection dominates over bottom and random, and combining Top activations with the Max mask criterion (highlighted) yields the best joint score in nearly all cases, confirming that high-importance activations carry the signal most relevant to subject disentanglement, and that a precise mask is essential for a balanced transport.
\begin{table}[t]
  \centering
  \setlength{\tabcolsep}{0.4em}
  \caption{Additional comparison in prompt-to-prompt semantic transport with additional baselines.}
  \vspace{-0.1cm}
  \resizebox{\linewidth}{!}{%
  \begin{tabular}{c l c c c c c c c c c c c c c c}
   \toprule
      &  &  &  & &  \multicolumn{3}{c}{\textbf{CLIP-T}} & &  \multicolumn{3}{c}{\textbf{CLIP-I}}  & &  \multicolumn{3}{c}{\textbf{DINO-I}} \\
     \cmidrule{6-8} \cmidrule{10-12} \cmidrule{14-16}
     &  & \textbf{Mask} & \textbf{Layers / Steps}& & \textbf{S*T} & \textbf{S} & \textbf{T} & &  \textbf{S*T} & \textbf{S} & \textbf{T} & &  \textbf{S*T} & \textbf{S} & \textbf{T} \\
    \midrule    
\multirow{3}{*}{FLUX.1-schnell}
     & Interpolation & - & all & & 31.6 & \textcolor{tablegray}{83.2} & \textcolor{tablegray}{38.0} & & 53.4 & \textcolor{tablegray}{100.0} & \textcolor{tablegray}{53.4} & & 2.7 & \textcolor{tablegray}{99.9} & \textcolor{tablegray}{2.7} \\
     & Interpolation & - & mid & & 31.6 & \textcolor{tablegray}{83.2} & \textcolor{tablegray}{38.0} & & 53.3 & \textcolor{tablegray}{99.8} & \textcolor{tablegray}{53.4} & & 2.7 & \textcolor{tablegray}{99.7} & \textcolor{tablegray}{2.7} \\
     & \cellcolor{OurColor}\textbf{MAs (Ours)} & \cellcolor{OurColor}\cmark & \cellcolor{OurColor}mid & \cellcolor{OurColor}& \cellcolor{OurColor}\textbf{44.1} & \cellcolor{OurColor}\textcolor{tablegray}{62.8} & \cellcolor{OurColor}\textcolor{tablegray}{70.3} & \cellcolor{OurColor}&  \cellcolor{OurColor}\textbf{55.2} & \cellcolor{OurColor}\textcolor{tablegray}{71.2} & \cellcolor{OurColor}\textcolor{tablegray}{77.6} & \cellcolor{OurColor}&  \cellcolor{OurColor}\textbf{20.1} & \cellcolor{OurColor}\textcolor{tablegray}{42.4} & \cellcolor{OurColor}\textcolor{tablegray}{47.5} \\
     \midrule
\multirow{5}{*}{FLUX.1-dev}
     & Interpolation & - & all & & 32.4 & \textcolor{tablegray}{82.0} & \textcolor{tablegray}{39.5} & & 47.8 & \textcolor{tablegray}{91.0} & \textcolor{tablegray}{52.5} & & 2.5 & \textcolor{tablegray}{86.2} & \textcolor{tablegray}{2.9} \\
     & Interpolation & - & mid & & 33.9 & \textcolor{tablegray}{82.9} & \textcolor{tablegray}{40.8} & & 45.4 & \textcolor{tablegray}{88.3} & \textcolor{tablegray}{51.5} & & 2.6 & \textcolor{tablegray}{84.7} & \textcolor{tablegray}{3.0} \\
     & Latent Interpolation & - & 1st & & 41.6 & \textcolor{tablegray}{63.1} & \textcolor{tablegray}{65.9} & & 55.2 & \textcolor{tablegray}{73.3} & \textcolor{tablegray}{75.3} & & \textbf{20.4} & \textcolor{tablegray}{42.3} & \textcolor{tablegray}{48.3} \\
     & Latent Interpolation & - & last & & 30.4 & \textcolor{tablegray}{37.7} & \textcolor{tablegray}{80.8} & & 53.9 & \textcolor{tablegray}{57.4} & \textcolor{tablegray}{93.9} & & 4.0 & \textcolor{tablegray}{4.8} & \textcolor{tablegray}{83.1} \\
     & \cellcolor{OurColor}\textbf{MAs (Ours)} & \cellcolor{OurColor}\cmark & \cellcolor{OurColor}mid & \cellcolor{OurColor}& \cellcolor{OurColor}\textbf{42.5} & \cellcolor{OurColor}\textcolor{tablegray}{69.2} & \cellcolor{OurColor}\textcolor{tablegray}{61.5} & \cellcolor{OurColor}&  \cellcolor{OurColor}\textbf{58.1} & \cellcolor{OurColor}\textcolor{tablegray}{82.6} & \cellcolor{OurColor}\textcolor{tablegray}{70.4} & \cellcolor{OurColor}&  \cellcolor{OurColor}19.2 & \cellcolor{OurColor}\textcolor{tablegray}{61.4} & \cellcolor{OurColor}\textcolor{tablegray}{31.3} \\
     \midrule
\multirow{5}{*}{FLUX.2-klein}
     & Interpolation & - & all & & 30.4 & \textcolor{tablegray}{81.2} & \textcolor{tablegray}{37.4} & & 52.8 & \textcolor{tablegray}{99.6} & \textcolor{tablegray}{53.0} & & 3.2 & \textcolor{tablegray}{99.5} & \textcolor{tablegray}{3.2} \\
     & Interpolation & - & mid & & 30.2 & \textcolor{tablegray}{81.1} & \textcolor{tablegray}{37.3} & & 52.9 & \textcolor{tablegray}{99.7} & \textcolor{tablegray}{53.1} & & 3.2 & \textcolor{tablegray}{99.5} & \textcolor{tablegray}{3.2} \\
     & Latent Interpolation & - & 1st & & 29.8 & \textcolor{tablegray}{37.2} & \textcolor{tablegray}{80.0} & & 50.4 & \textcolor{tablegray}{54.0} & \textcolor{tablegray}{93.4} & & 3.6 & \textcolor{tablegray}{4.2} & \textcolor{tablegray}{86.1} \\
     & Latent Interpolation & - & last & & 39.1 & \textcolor{tablegray}{61.2} & \textcolor{tablegray}{63.9} & & 47.7 & \textcolor{tablegray}{68.5} & \textcolor{tablegray}{69.6} & & 13.3 & \textcolor{tablegray}{33.4} & \textcolor{tablegray}{39.7} \\
     & \cellcolor{OurColor}\textbf{MAs (Ours)} & \cellcolor{OurColor}\cmark & \cellcolor{OurColor}mid & \cellcolor{OurColor}& \cellcolor{OurColor}\textbf{41.7} & \cellcolor{OurColor}\textcolor{tablegray}{64.6} & \cellcolor{OurColor}\textcolor{tablegray}{64.5} & \cellcolor{OurColor}& \cellcolor{OurColor}\textbf{55.2} & \cellcolor{OurColor}\textcolor{tablegray}{75.2} & \cellcolor{OurColor}\textcolor{tablegray}{73.4} & \cellcolor{OurColor}& \cellcolor{OurColor}\textbf{19.1} & \cellcolor{OurColor}\textcolor{tablegray}{58.0} & \cellcolor{OurColor}\textcolor{tablegray}{32.9}\\
     \midrule
\multirow{5}{*}{Qwen-Image}
     & Interpolation & - & all & & 22.1 & \textcolor{tablegray}{47.1} & \textcolor{tablegray}{47.0} & & 25.1 & \textcolor{tablegray}{50.6} & \textcolor{tablegray}{49.6} & & 0.0 & \textcolor{tablegray}{0.8} & \textcolor{tablegray}{0.7} \\
     & Interpolation & - & mid & & 22.8 & \textcolor{tablegray}{47.8} & \textcolor{tablegray}{47.7} & & 25.6 & \textcolor{tablegray}{51.1} & \textcolor{tablegray}{50.1} & & 0.0 & \textcolor{tablegray}{0.7} & \textcolor{tablegray}{0.6} \\
     & Latent Interpolation & - & 1st & & 42.0 & \textcolor{tablegray}{63.8} & \textcolor{tablegray}{65.8} & & 52.0 & \textcolor{tablegray}{71.7} & \textcolor{tablegray}{72.6} & & 15.9 & \textcolor{tablegray}{37.9} & \textcolor{tablegray}{41.9} \\
     & Latent Interpolation & - & last & & 30.2 & \textcolor{tablegray}{36.5} & \textcolor{tablegray}{82.8} & & 54.2 & \textcolor{tablegray}{55.8} & \textcolor{tablegray}{97.3} & & 3.5 & \textcolor{tablegray}{3.7} & \textcolor{tablegray}{93.1} \\
     & \cellcolor{OurColor}\textbf{MAs (Ours)} & \cellcolor{OurColor}\cmark &\cellcolor{OurColor}mid & \cellcolor{OurColor}& \cellcolor{OurColor}\textbf{44.5} & \cellcolor{OurColor}\textcolor{tablegray}{67.0} & \cellcolor{OurColor}\textcolor{tablegray}{66.4} & \cellcolor{OurColor}&  \cellcolor{OurColor}55.4 & \cellcolor{OurColor}\textcolor{tablegray}{76.0} & \cellcolor{OurColor}\textcolor{tablegray}{72.9} & \cellcolor{OurColor}&  \cellcolor{OurColor}\textbf{18.0} & \cellcolor{OurColor}\textcolor{tablegray}{48.6} & \cellcolor{OurColor}\textcolor{tablegray}{37.1} \\
     \midrule
\multirow{5}{*}{SANA1.5}
     & Interpolation & - & all & & 31.4 & \textcolor{tablegray}{66.3} & \textcolor{tablegray}{47.3} & & 34.8 & \textcolor{tablegray}{64.9} & \textcolor{tablegray}{53.7} & & 1.4 & \textcolor{tablegray}{34.1} & \textcolor{tablegray}{4.2} \\
     & Interpolation & - & mid & & 32.6 & \textcolor{tablegray}{69.3} & \textcolor{tablegray}{47.0} & & 36.8 & \textcolor{tablegray}{67.9} & \textcolor{tablegray}{54.3} & & 2.0 & \textcolor{tablegray}{43.4} & \textcolor{tablegray}{4.6} \\
     & Latent Interpolation & - & 1st & & 31.4 & \textcolor{tablegray}{37.9} & \textcolor{tablegray}{82.9} & & 55.0 & \textcolor{tablegray}{56.7} & \textcolor{tablegray}{96.9} & & 5.1 & \textcolor{tablegray}{5.4} & \textcolor{tablegray}{93.1} \\
     & Latent Interpolation & - & last & & 40.2 & \textcolor{tablegray}{62.2} & \textcolor{tablegray}{64.6} & & 53.2 & \textcolor{tablegray}{72.2} & \textcolor{tablegray}{73.7} & & 14.9 & \textcolor{tablegray}{35.7} & \textcolor{tablegray}{41.8} \\
     & \cellcolor{OurColor}\textbf{MAs (Ours)} & \cellcolor{OurColor}\cmark & \cellcolor{OurColor}mid & \cellcolor{OurColor}& \cellcolor{OurColor}\textbf{41.4} & \cellcolor{OurColor}\textcolor{tablegray}{72.8} & \cellcolor{OurColor}\textcolor{tablegray}{56.8} & \cellcolor{OurColor}&  \cellcolor{OurColor}\textbf{55.7} & \cellcolor{OurColor}\textcolor{tablegray}{83.7} & \cellcolor{OurColor}\textcolor{tablegray}{66.6} & \cellcolor{OurColor}&  \cellcolor{OurColor}\textbf{17.0} & \cellcolor{OurColor}\textcolor{tablegray}{65.0} & \cellcolor{OurColor}\textcolor{tablegray}{26.1} \\
    \bottomrule
  \end{tabular}%
  }
  \label{tab:sup_prompt_prompt_latent}
  \vspace{-0.35cm}
\end{table}
\begin{table}[t]
  \centering
  \setlength{\tabcolsep}{0.75em}
  \caption{Prompt-to-prompt composition under different activation (random, bottom-$k$, top-$k$) and masking strategies. Scores are reported for source (S), target (B), and joint preservation (S*T) using CLIP-T, CLIP-I, and DINO similarity.}
  \vspace{-0.1cm}
  \resizebox{0.95\linewidth}{!}{%
  \begin{tabular}{c l l c c c c c c c c c c c c}
    \toprule
      & \multicolumn{2}{c}{\textbf{Criterion}} & &  \multicolumn{3}{c}{\textbf{CLIP-T}} & &  \multicolumn{3}{c}{\textbf{CLIP-I}}  & &  \multicolumn{3}{c}{\textbf{DINO}} \\
     \cmidrule{2-3} \cmidrule{5-7} \cmidrule{9-11} \cmidrule{13-15}
     & \textbf{Activations} & \textbf{Mask}& &  \textbf{S*T} & \textbf{S} & \textbf{T} & &  \textbf{S*T} & \textbf{S} & \textbf{T} & &  \textbf{S*T} & \textbf{S} & \textbf{T} \\
    \midrule
    \multirow{9}{*}{\rotatebox{90}{\makecell{FLUX.1-schnell}}}
     & \multirow{3}{*}{Random} & Random & & 34.6 & \textcolor{tablegray}{42.3} & \textcolor{tablegray}{81.8} & &  52.2 & \textcolor{tablegray}{56.5} & \textcolor{tablegray}{92.4} & &  6.9 & \textcolor{tablegray}{8.4} & \textcolor{tablegray}{82.3} \\
     &  & Min & & 34.9 & \textcolor{tablegray}{42.9} & \textcolor{tablegray}{81.5} & &  52.4 & \textcolor{tablegray}{57.0} & \textcolor{tablegray}{91.9} & &  7.1 & \textcolor{tablegray}{8.7} & \textcolor{tablegray}{81.5} \\
     &  & Max & & 39.0 & \textcolor{tablegray}{50.1} & \textcolor{tablegray}{77.9} & &  53.3 & \textcolor{tablegray}{61.3} & \textcolor{tablegray}{87.0} & &  14.8 & \textcolor{tablegray}{22.6} & \textcolor{tablegray}{65.5} \\
     \cmidrule{2-15}
     & \multirow{3}{*}{Bottom} & Random & & 38.1 & \textcolor{tablegray}{48.1} & \textcolor{tablegray}{79.2} & &  53.0 & \textcolor{tablegray}{60.0} & \textcolor{tablegray}{88.4} & &  11.9 & \textcolor{tablegray}{16.4} & \textcolor{tablegray}{72.6} \\
     &  & Min & & 38.9 & \textcolor{tablegray}{49.5} & \textcolor{tablegray}{78.6} & &  53.3 & \textcolor{tablegray}{61.0} & \textcolor{tablegray}{87.4} & &  12.8 & \textcolor{tablegray}{18.1} & \textcolor{tablegray}{70.4} \\
     &  & Max & & 42.3 & \textcolor{tablegray}{57.4} & \textcolor{tablegray}{73.8} & &  54.5 & \textcolor{tablegray}{66.6} & \textcolor{tablegray}{81.7} & & 18.6 & \textcolor{tablegray}{34.0} & \textcolor{tablegray}{54.8} \\
     \cmidrule{2-15}
     & \multirow{3}{*}{Top} & Random & & 42.1 & \textcolor{tablegray}{55.7} & \textcolor{tablegray}{75.6} & &  53.9 & \textcolor{tablegray}{65.0} & \textcolor{tablegray}{82.8} & &  15.6 & \textcolor{tablegray}{25.1} & \textcolor{tablegray}{62.2} \\
     &  & Min & & 42.4 & \textcolor{tablegray}{56.8} & \textcolor{tablegray}{74.8} & &  54.0 & \textcolor{tablegray}{65.9} & \textcolor{tablegray}{81.9} & &  16.1 & \textcolor{tablegray}{26.5} & \textcolor{tablegray}{60.7} \\
     &  & \cellcolor{OurColor}Max & \cellcolor{OurColor}& \cellcolor{OurColor}\textbf{44.1} & \cellcolor{OurColor}\textcolor{tablegray}{62.8} & \cellcolor{OurColor}\textcolor{tablegray}{70.2} & \cellcolor{OurColor}&  \cellcolor{OurColor}\textbf{55.2} & \cellcolor{OurColor}\textcolor{tablegray}{71.2} & \cellcolor{OurColor}\textcolor{tablegray}{77.6} & \cellcolor{OurColor}& \cellcolor{OurColor} \textbf{20.1} & \cellcolor{OurColor}\textcolor{tablegray}{42.4} & \cellcolor{OurColor}\textcolor{tablegray}{47.5} \\
     \midrule
    \multirow{9}{*}{\rotatebox{90}{\makecell{FLUX.1-dev}}}
     & \multirow{3}{*}{Random} & Random & & 40.2 & \textcolor{tablegray}{55.1} & \textcolor{tablegray}{73.1} & &  55.4 & \textcolor{tablegray}{68.8} & \textcolor{tablegray}{80.5} & &  14.8 & \textcolor{tablegray}{27.4} & \textcolor{tablegray}{54.0} \\
     & & Min & & 37.4 & \textcolor{tablegray}{48.9} & \textcolor{tablegray}{76.5} & &  54.5 & \textcolor{tablegray}{64.5} & \textcolor{tablegray}{84.6} & &  10.0 & \textcolor{tablegray}{15.5} & \textcolor{tablegray}{64.7} \\
     & & Max & & 40.9 & \textcolor{tablegray}{57.0} & \textcolor{tablegray}{71.8} & &  56.6 & \textcolor{tablegray}{70.2} & \textcolor{tablegray}{80.6} & &  18.5 & \textcolor{tablegray}{36.7} & \textcolor{tablegray}{50.5} \\
     \cmidrule{2-15}
     & \multirow{3}{*}{Bottom} & Random & & 37.7 & \textcolor{tablegray}{50.0} & \textcolor{tablegray}{75.3} & &  55.9 & \textcolor{tablegray}{66.3} & \textcolor{tablegray}{84.4} & &  14.0 & \textcolor{tablegray}{23.1} & \textcolor{tablegray}{60.4} \\
     & & Min & & 34.5 & \textcolor{tablegray}{44.1} & \textcolor{tablegray}{78.1} & &  55.7 & \textcolor{tablegray}{62.6} & \textcolor{tablegray}{89.0} & &  9.3 & \textcolor{tablegray}{13.0} & \textcolor{tablegray}{71.7} \\
     & & Max & & 38.8 & \textcolor{tablegray}{52.4} & \textcolor{tablegray}{74.1} & &  56.9 & \textcolor{tablegray}{67.5} & \textcolor{tablegray}{84.2} & &  17.4 & \textcolor{tablegray}{30.3} & \textcolor{tablegray}{57.6} \\
     \cmidrule{2-15}
     & \multirow{3}{*}{Top} & Random & & 40.7 & \textcolor{tablegray}{71.4} & \textcolor{tablegray}{57.0} & &  56.8 & \textcolor{tablegray}{85.4} & \textcolor{tablegray}{66.5} & &  16.0 & \textcolor{tablegray}{62.1} & \textcolor{tablegray}{25.8} \\
     & & Min & & 39.0 & \textcolor{tablegray}{74.0} & \textcolor{tablegray}{52.7} & &  56.3 & \textcolor{tablegray}{88.6} & \textcolor{tablegray}{63.5} & &  14.6 & \textcolor{tablegray}{68.0} & \textcolor{tablegray}{21.5} \\
     & & \cellcolor{OurColor}Max & \cellcolor{OurColor}& \cellcolor{OurColor}\textbf{42.5} & \cellcolor{OurColor}\textcolor{tablegray}{69.2} & \cellcolor{OurColor}\textcolor{tablegray}{61.5} & \cellcolor{OurColor}&  \cellcolor{OurColor}\textbf{58.1} & \cellcolor{OurColor}\textcolor{tablegray}{82.6} & \cellcolor{OurColor}\textcolor{tablegray}{70.4} & \cellcolor{OurColor}&  \cellcolor{OurColor}\textbf{19.2} & \cellcolor{OurColor}\textcolor{tablegray}{61.4} & \cellcolor{OurColor}\textcolor{tablegray}{31.3} \\
     \midrule
    \multirow{9}{*}{\rotatebox{90}{\makecell{FLUX.2-klein}}}
     & \multirow{3}{*}{Random} & Random & & 40.4 & \textcolor{tablegray}{59.8} & \textcolor{tablegray}{67.6} & &  53.7 & \textcolor{tablegray}{70.5} & \textcolor{tablegray}{76.1} & &  17.7 & \textcolor{tablegray}{39.5} & \textcolor{tablegray}{44.8} \\
     & & Min & & 40.6 & \textcolor{tablegray}{60.7} & \textcolor{tablegray}{66.9} & &  53.5 & \textcolor{tablegray}{71.5} & \textcolor{tablegray}{74.9} & &  17.6 & \textcolor{tablegray}{41.7} & \textcolor{tablegray}{42.2} \\
     & & Max & & 40.9 & \textcolor{tablegray}{61.1} & \textcolor{tablegray}{66.9} & &  54.5 & \textcolor{tablegray}{71.6} & \textcolor{tablegray}{76.1} & &  19.1 & \textcolor{tablegray}{50.7} & \textcolor{tablegray}{37.7} \\
     \cmidrule{2-15}
     & \multirow{3}{*}{Bottom} & Random & & 40.3 & \textcolor{tablegray}{61.2} & \textcolor{tablegray}{65.9} & &  53.3 & \textcolor{tablegray}{71.7} & \textcolor{tablegray}{74.3} & &  16.6 & \textcolor{tablegray}{37.5} & \textcolor{tablegray}{44.4} \\
     & & Min & & 40.4 & \textcolor{tablegray}{61.5} & \textcolor{tablegray}{65.6} & &  52.8 & \textcolor{tablegray}{71.9} & \textcolor{tablegray}{73.5} & &  16.2 & \textcolor{tablegray}{37.1} & \textcolor{tablegray}{43.8} \\
     & & Max & & 41.2 & \textcolor{tablegray}{63.2} & \textcolor{tablegray}{65.3} & &  54.9 & \textcolor{tablegray}{73.5} & \textcolor{tablegray}{74.7} & &  \textbf{19.2} & \textcolor{tablegray}{51.2} & \textcolor{tablegray}{37.5} \\
     \cmidrule{2-15}
     & \multirow{3}{*}{Top} & Random & & 41.2 & \textcolor{tablegray}{63.7} & \textcolor{tablegray}{64.7} & &  54.4 & \textcolor{tablegray}{74.5} & \textcolor{tablegray}{73.0} & &  18.7 & \textcolor{tablegray}{48.1} & \textcolor{tablegray}{38.9} \\
     &  & Min & & 41.5 & \textcolor{tablegray}{65.6} & \textcolor{tablegray}{63.3} & & 54.2 & \textcolor{tablegray}{76.3} & \textcolor{tablegray}{71.0} & &  18.2 & \textcolor{tablegray}{52.0} & \textcolor{tablegray}{34.9} \\
     &  & \cellcolor{OurColor}Max & \cellcolor{OurColor}& \cellcolor{OurColor}\textbf{41.7} & \cellcolor{OurColor}\textcolor{tablegray}{64.6} & \cellcolor{OurColor}\textcolor{tablegray}{64.5} & \cellcolor{OurColor}&  \cellcolor{OurColor}\textbf{55.2} & \cellcolor{OurColor}\textcolor{tablegray}{75.2} & \cellcolor{OurColor}\textcolor{tablegray}{73.4} & \cellcolor{OurColor}&  \cellcolor{OurColor}19.1 & \cellcolor{OurColor}\textcolor{tablegray}{58.0} & \cellcolor{OurColor}\textcolor{tablegray}{32.9} \\
     \midrule
    \multirow{9}{*}{\rotatebox{90}{\makecell{Qwen-Image}}}
     & \multirow{3}{*}{Random} & Random & & 32.4 & \textcolor{tablegray}{39.0} & \textcolor{tablegray}{82.9} & &  52.3 & \textcolor{tablegray}{56.2} & \textcolor{tablegray}{93.0} & &  3.5 & \textcolor{tablegray}{4.1} & \textcolor{tablegray}{84.5} \\
     &  & Min & & 32.3 & \textcolor{tablegray}{38.9} & \textcolor{tablegray}{82.8} & &  52.2 & \textcolor{tablegray}{56.1} & \textcolor{tablegray}{93.2} & &  3.5 & \textcolor{tablegray}{4.1} & \textcolor{tablegray}{84.2} \\
     &  & Max & & 32.8 & \textcolor{tablegray}{39.7} & \textcolor{tablegray}{82.6} & &  52.1 & \textcolor{tablegray}{56.5} & \textcolor{tablegray}{92.2} & &  4.0 & \textcolor{tablegray}{4.9} & \textcolor{tablegray}{81.1} \\
     \cmidrule{2-15}
     & \multirow{3}{*}{Bottom}& Random & & 30.9 & \textcolor{tablegray}{37.3} & \textcolor{tablegray}{82.8} & &  53.9 & \textcolor{tablegray}{56.3} & \textcolor{tablegray}{95.7} & &  3.8 & \textcolor{tablegray}{4.3} & \textcolor{tablegray}{89.2} \\
     &  & Min & & 30.9 & \textcolor{tablegray}{37.4} & \textcolor{tablegray}{82.8} & &  53.9 & \textcolor{tablegray}{56.3} & \textcolor{tablegray}{95.7} & &  3.9 & \textcolor{tablegray}{4.4} & \textcolor{tablegray}{88.9} \\
     &  & Max & & 31.5 & \textcolor{tablegray}{38.2} & \textcolor{tablegray}{82.5} & &  53.5 & \textcolor{tablegray}{56.6} & \textcolor{tablegray}{94.4} & & 4.5 & \textcolor{tablegray}{5.2} & \textcolor{tablegray}{85.3} \\
     \cmidrule{2-15}
     & \multirow{3}{*}{Top} & Random & & 43.3 & \textcolor{tablegray}{66.0} & \textcolor{tablegray}{65.7} & & 52.1 & \textcolor{tablegray}{73.7} & \textcolor{tablegray}{70.7} & &  14.9 & \textcolor{tablegray}{40.1} & \textcolor{tablegray}{37.1} \\
     &  & Min & & 44.1 & \textcolor{tablegray}{65.4} & \textcolor{tablegray}{67.5} & &  53.8 & \textcolor{tablegray}{73.6} & \textcolor{tablegray}{73.1} & &  16.7 & \textcolor{tablegray}{42.1} & \textcolor{tablegray}{39.7} \\
     &  & \cellcolor{OurColor}Max & \cellcolor{OurColor}& \cellcolor{OurColor}\textbf{44.5} & \cellcolor{OurColor}\textcolor{tablegray}{67.0} & \cellcolor{OurColor}\textcolor{tablegray}{66.4} & \cellcolor{OurColor}&  \cellcolor{OurColor}\textbf{55.4} & \cellcolor{OurColor}\textcolor{tablegray}{76.0} & \cellcolor{OurColor}\textcolor{tablegray}{72.9} & \cellcolor{OurColor}&  \cellcolor{OurColor}\textbf{18.0} & \cellcolor{OurColor}\textcolor{tablegray}{48.6} & \cellcolor{OurColor}\textcolor{tablegray}{37.1} \\
     \midrule
    \multirow{9}{*}{\rotatebox{90}{\makecell{SANA1.5}}}
     & \multirow{3}{*}{Random} & Random & & 39.8 & \textcolor{tablegray}{73.1} & \textcolor{tablegray}{54.4} & &  53.8 & \textcolor{tablegray}{83.4} & \textcolor{tablegray}{64.5} & &  14.9 & \textcolor{tablegray}{56.6} & \textcolor{tablegray}{26.4} \\
     &  & Min & & 40.2 & \textcolor{tablegray}{71.3} & \textcolor{tablegray}{56.4} & &  53.6 & \textcolor{tablegray}{81.2} & \textcolor{tablegray}{66.0} & &  15.3 & \textcolor{tablegray}{50.5} & \textcolor{tablegray}{30.4} \\
     &  & Max & & 40.7 & \textcolor{tablegray}{73.1} & \textcolor{tablegray}{55.6} & &  55.4 & \textcolor{tablegray}{83.8} & \textcolor{tablegray}{66.1} & &  \textbf{16.0} & \textcolor{tablegray}{65.1} & \textcolor{tablegray}{24.6} \\
     \cmidrule{2-15}
     & \multirow{3}{*}{Bottom} & Random & & 38.9 & \textcolor{tablegray}{71.3} & \textcolor{tablegray}{54.5} & &  52.3 & \textcolor{tablegray}{80.7} & \textcolor{tablegray}{64.8} & &  13.6 & \textcolor{tablegray}{48.2} & \textcolor{tablegray}{28.1} \\
     &  & Min & & 39.6 & \textcolor{tablegray}{70.6} & \textcolor{tablegray}{56.2} & &  52.7 & \textcolor{tablegray}{80.0} & \textcolor{tablegray}{65.9} & &  14.6 & \textcolor{tablegray}{47.6} & \textcolor{tablegray}{30.6} \\
     &  & Max & & 40.2 & \textcolor{tablegray}{72.5} & \textcolor{tablegray}{55.5} & &  54.6 & \textcolor{tablegray}{82.8} & \textcolor{tablegray}{66.0} & &  15.7 & \textcolor{tablegray}{61.8} & \textcolor{tablegray}{25.4} \\
     \cmidrule{2-15}
     & \multirow{3}{*}{Top} & Random & & 39.8 & \textcolor{tablegray}{74.6} & \textcolor{tablegray}{53.4} & &  54.4 & \textcolor{tablegray}{85.2} & \textcolor{tablegray}{63.8} & &  14.7 & \textcolor{tablegray}{61.2} & \textcolor{tablegray}{24.1} \\
     &  & Min & & 40.7 & \textcolor{tablegray}{73.7} & \textcolor{tablegray}{55.1} & &  54.7 & \textcolor{tablegray}{84.3} & \textcolor{tablegray}{65.0} & &  15.9 & \textcolor{tablegray}{59.4} & \textcolor{tablegray}{26.8} \\
     &  & \cellcolor{OurColor}Max &\cellcolor{OurColor} & \cellcolor{OurColor}\textbf{41.0} & \cellcolor{OurColor}\textcolor{tablegray}{74.9} & \cellcolor{OurColor}\textcolor{tablegray}{54.7} & \cellcolor{OurColor}&  \cellcolor{OurColor}\textbf{56.2} & \cellcolor{OurColor}\textcolor{tablegray}{86.4} & \cellcolor{OurColor}\textcolor{tablegray}{65.1} & \cellcolor{OurColor}&  \cellcolor{OurColor}15.7 & \cellcolor{OurColor}\textcolor{tablegray}{71.3} & \cellcolor{OurColor}\textcolor{tablegray}{22.1} \\
    \bottomrule
  \end{tabular}%
  }
  \label{tab:sup_prompt_prompt_top_k}
  \vspace{-0.4cm}
\end{table}
\begin{table}[h]
\setlength{\tabcolsep}{1em}
\caption{Per-model preference rate (\%) of MAs vs TokenVerse 
across the three Likert axes. Ties are split 50/50. Bold marks the preferred side.}
\centering
\resizebox{0.9\linewidth}{!}{%
\begin{tabular}{l cc cc cc}
\toprule
 & \multicolumn{2}{c}{Prompt Alignment} & \multicolumn{2}{c}{Subject Faithfulness} & \multicolumn{2}{c}{Visual Quality} \\
\cmidrule(lr){2-3} \cmidrule(lr){4-5} \cmidrule(lr){6-7}
Model & TokenVerse & \cellcolor{OurColor}\textbf{MAs} & TokenVerse & \cellcolor{OurColor}\textbf{MAs} & TokenVerse & \cellcolor{OurColor}\textbf{MAs} \\
\midrule
FLUX.1-schnell  & 36.7 & \cellcolor{OurColor}\textbf{63.3} & 31.6 & \cellcolor{OurColor}\textbf{68.4} & 21.4 & \cellcolor{OurColor}\textbf{78.6} \\
FLUX.1-dev      & 35.6 & \cellcolor{OurColor}\textbf{64.4} & 7.8  & \cellcolor{OurColor}\textbf{92.2} & 31.1 & \cellcolor{OurColor}\textbf{68.9} \\
FLUX.2-klein  & 29.0 & \cellcolor{OurColor}\textbf{71.0} & 33.6 & \cellcolor{OurColor}\textbf{66.4} & 26.4 & \cellcolor{OurColor}\textbf{73.6} \\
Qwen-Image      & 31.7 & \cellcolor{OurColor}\textbf{68.3} & 25.6 & \cellcolor{OurColor}\textbf{74.4} & 23.2 & \cellcolor{OurColor}\textbf{76.8} \\
SANA1.5         & 40.3 & \cellcolor{OurColor}\textbf{59.7} & 42.8 & \cellcolor{OurColor}\textbf{57.2} & 41.4 & \cellcolor{OurColor}\textbf{58.6} \\
\bottomrule
\end{tabular}
}
\label{tab:user-study-results}
\vspace{-0.3cm}
\end{table}

\begin{figure}[t]
    \centering
    \includegraphics[width=0.95\linewidth]{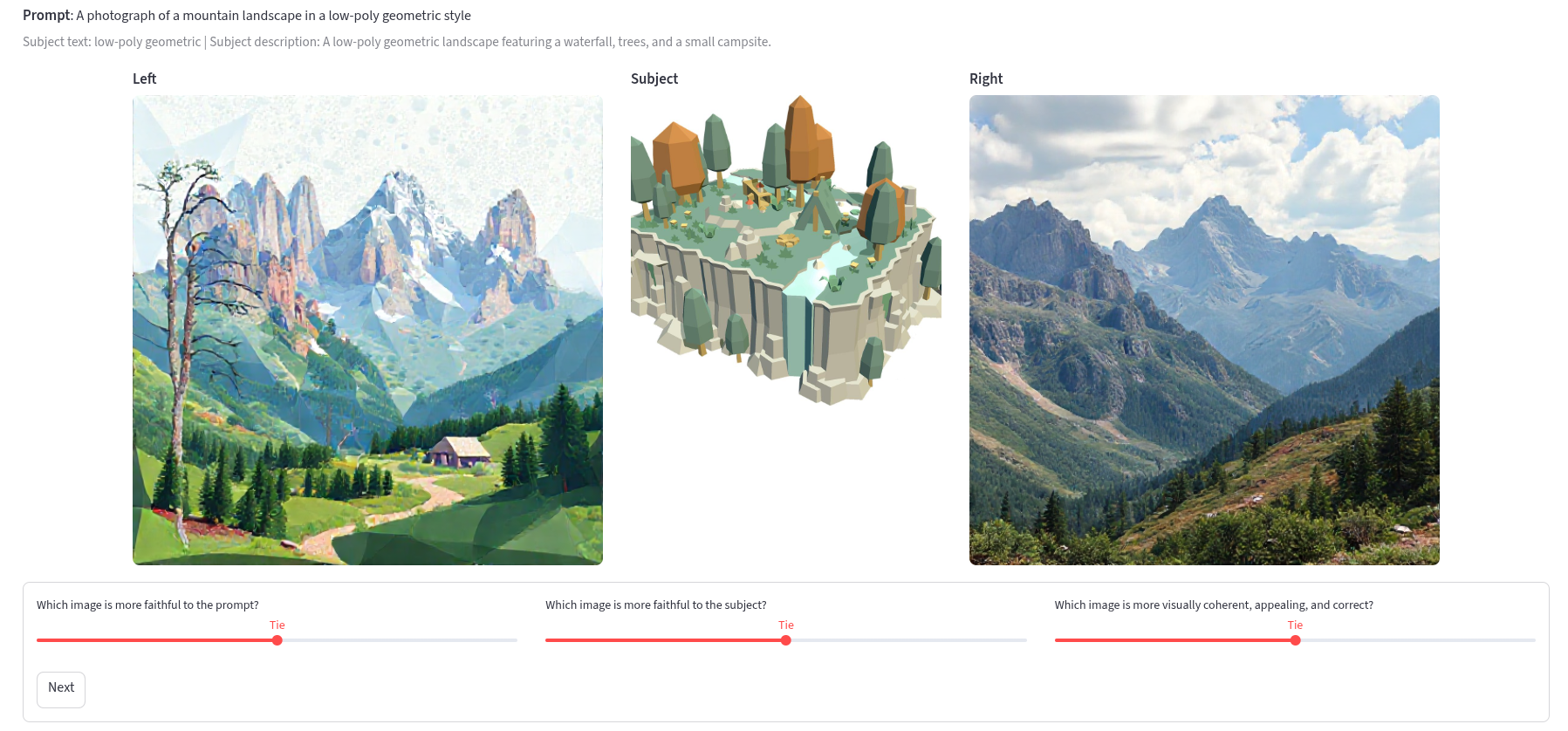}
    \vspace{-0.25cm}
    \caption{Screenshot of the platform used for collecting user-study data.}
    \label{fig:supp_user_study}
    \vspace{-0.3cm}
\end{figure}

\subsection{User Study on Image-Conditioned Transport}
We run a user study on DreamBench++ examples, sampling $150$ subjects-prompt pairs per backbone and collecting three independent preference judgments per sample between MAs and TokenVerse. Annotators rate each pair on a Likert scale that admits ties, along three axes: prompt alignment, subject faithfulness and general visual quality. Ties are split evenly between the two methods when computing preference rates. Table~\ref{tab:user-study-results} summarizes the outcome of the study. MAs are preferred over TokenVerse across all backbones and along all three evaluated axes, namely prompt alignment, subject faithfulness and visual quality. The preference is consistent regardless of the underlying generator, suggesting that the gains brought by MAs reflect a general utility in image-conditioned transport rather than an effect tied to a specific architecture.

\section{Additional Qualitative Results}
\begin{figure}[t]
    \centering    \includegraphics[width=0.98\linewidth]{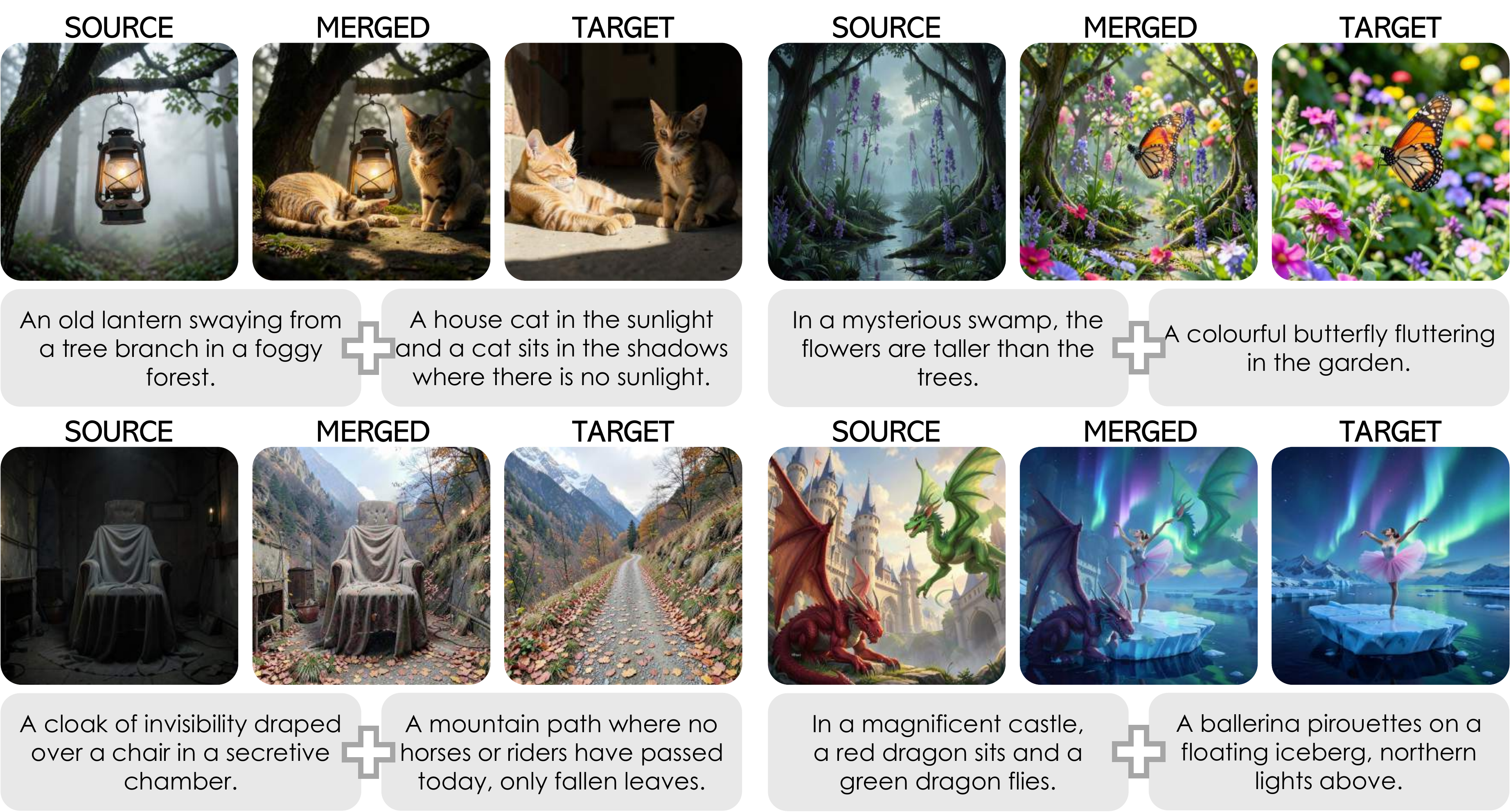}
    \vspace{-0.2cm}
    \caption{Qualitative examples of MAs-based text-conditioned transport on FLUX.1-dev.}
    \label{fig:prompt_prompt_dev}
    \vspace{-0.35cm}
\end{figure}

\begin{figure}[t]
    \centering    \includegraphics[width=0.98\linewidth]{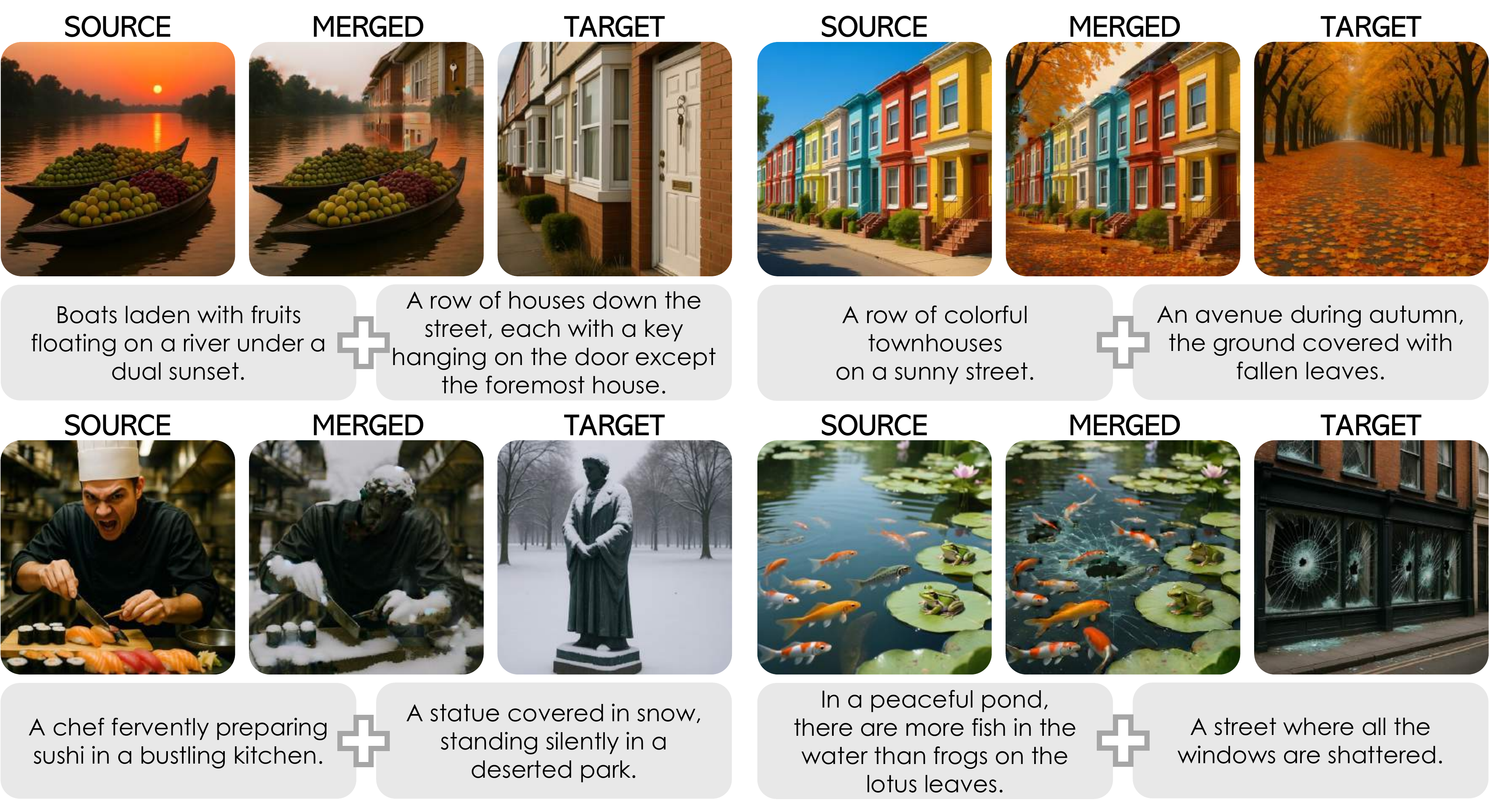}
    \vspace{-0.3cm}
    \caption{Qualitative examples of MAs-based text-conditioned transport on Qwen-Image.}
    \label{fig:prompt_prompt_qwen}
    \vspace{-0.35cm}
\end{figure}

\begin{figure}[t]
    \centering    \includegraphics[width=0.98\linewidth]{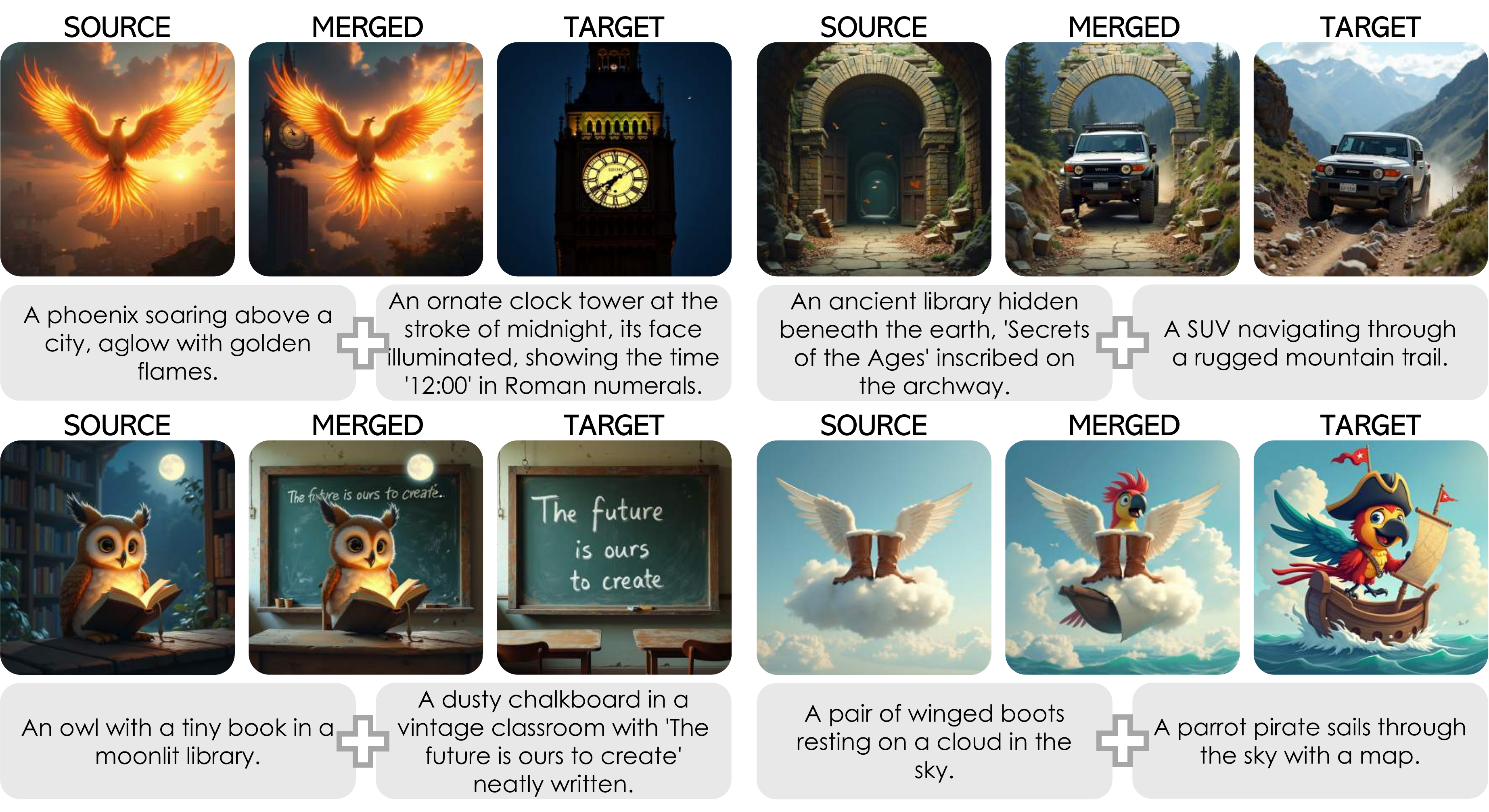}
    \vspace{-0.3cm}
    \caption{Qualitative examples of MAs-based text-conditioned transport on FLUX.2-klein.}
    \label{fig:prompt_prompt_klein}
    \vspace{-0.35cm}
\end{figure}

\begin{figure}[t]
    \centering    \includegraphics[width=0.98\linewidth]{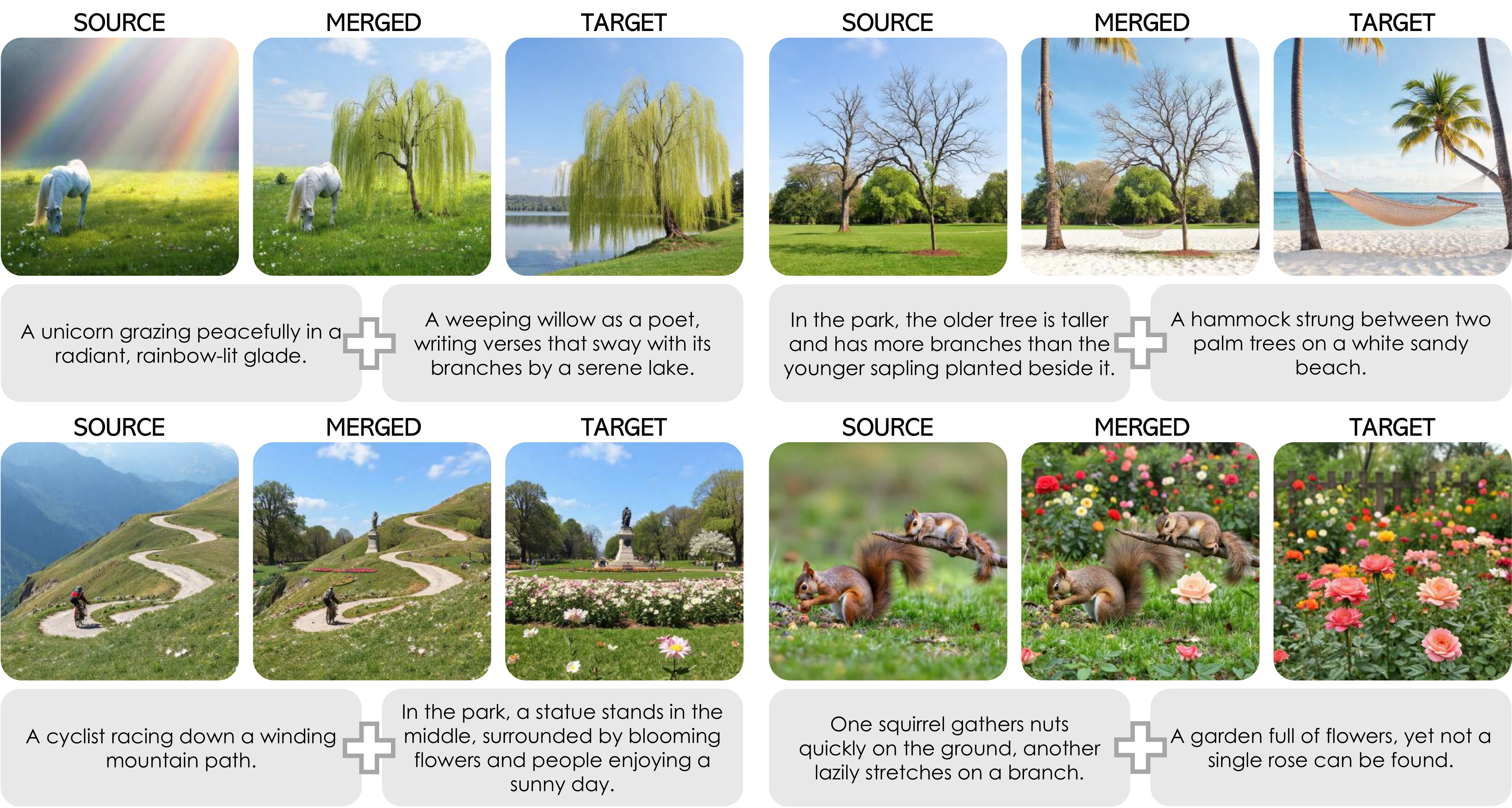}
    \vspace{-0.3cm}
    \caption{Qualitative examples of MAs-based text-conditioned transport on FLUX.1-schnell.}
    \label{fig:prompt_prompt_schnell}
    \vspace{-0.35cm}
\end{figure}

\begin{figure}[t]
    \centering    \includegraphics[width=0.98\linewidth]{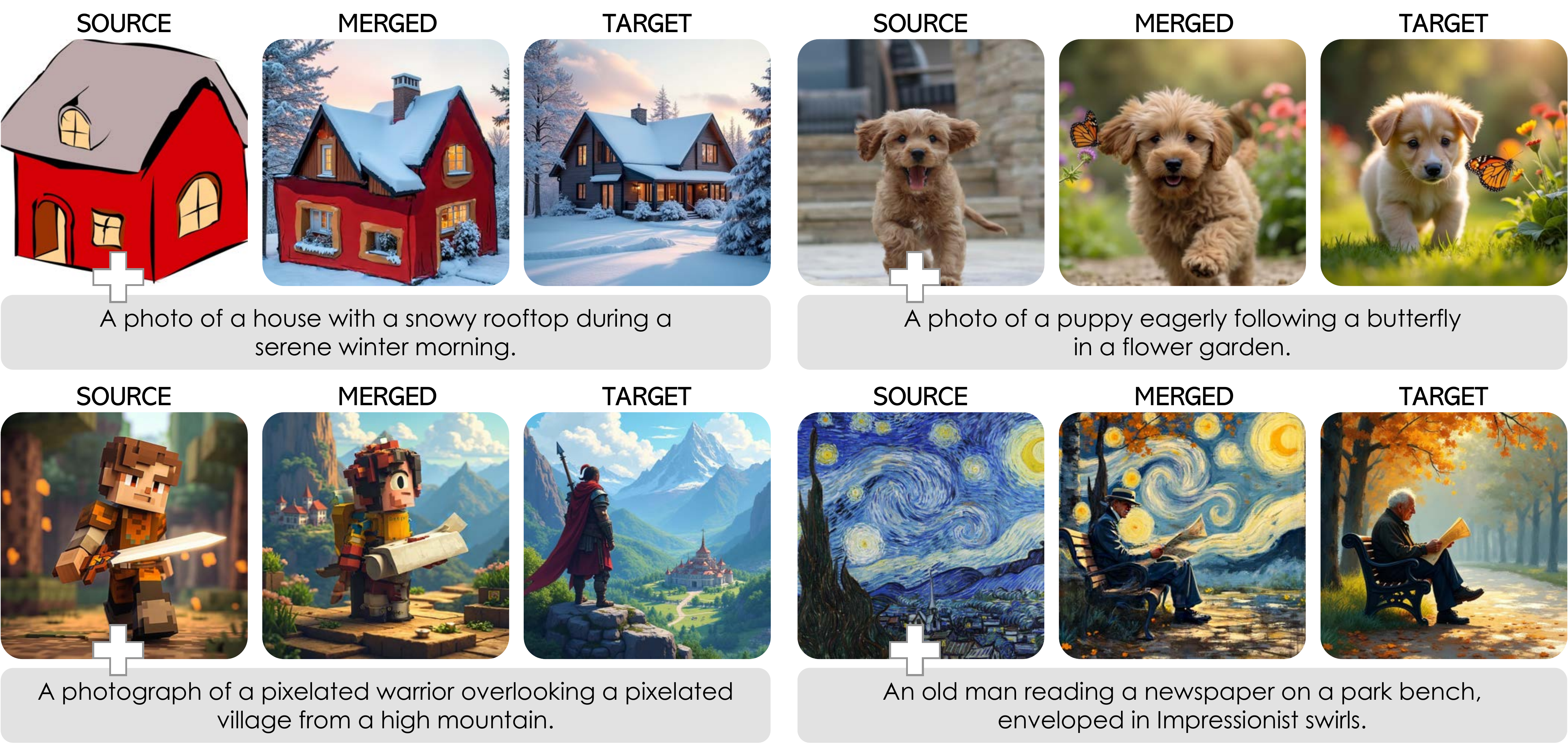}
    \vspace{-0.2cm}
    \caption{Qualitative examples of MAs-based image-conditioned transport on FLUX.1-dev.}
    \label{fig:img_prompt_dev}
    \vspace{-0.3cm}
\end{figure}

\begin{figure}[t]
    \centering    \includegraphics[width=0.98\linewidth]{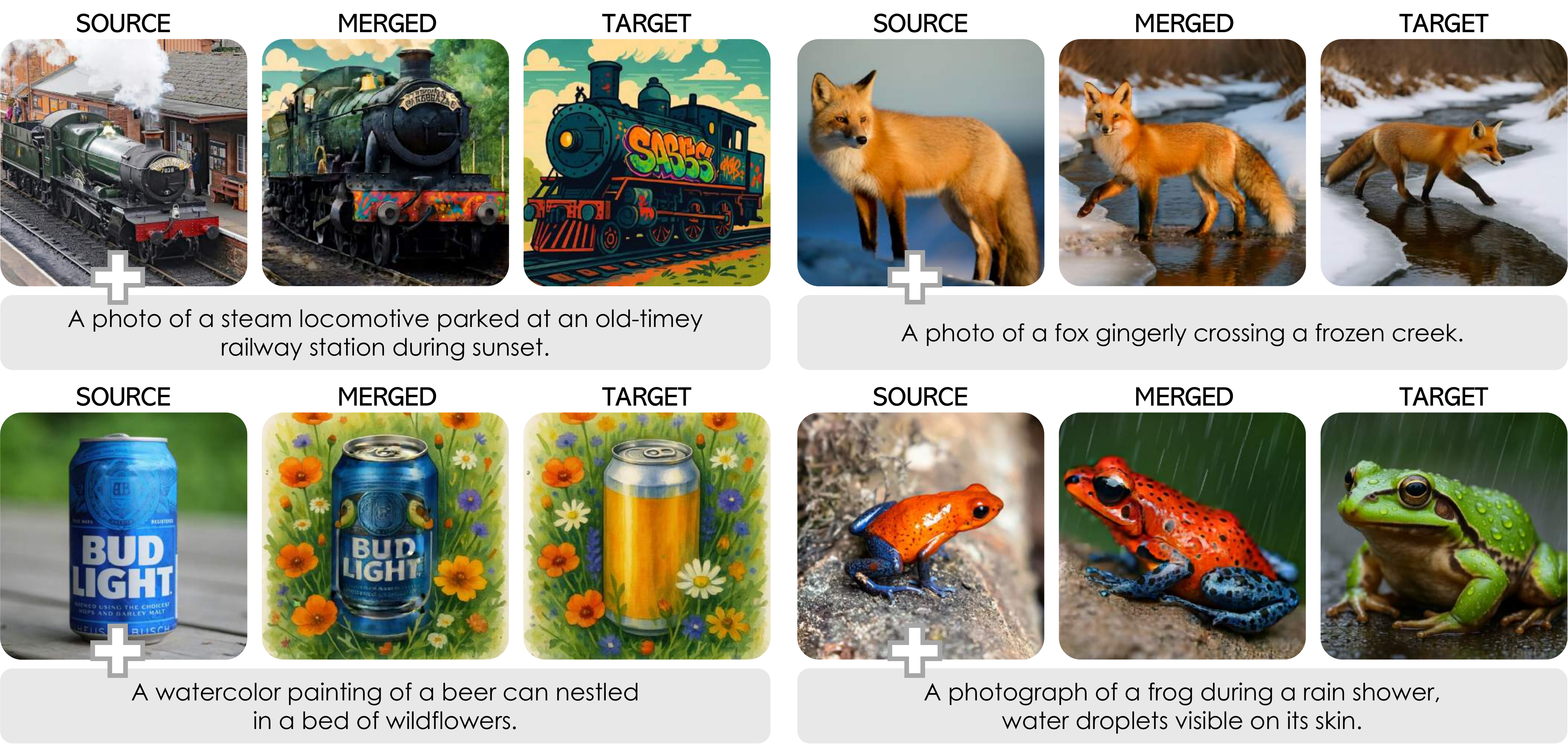}
    \vspace{-0.2cm}
    \caption{Qualitative examples of MAs-based image-conditioned transport on Qwen-Image.}
    \label{fig:img_prompt_qwen}
    \vspace{-0.3cm}
\end{figure}

\begin{figure}[t]
    \centering    \includegraphics[width=0.98\linewidth]{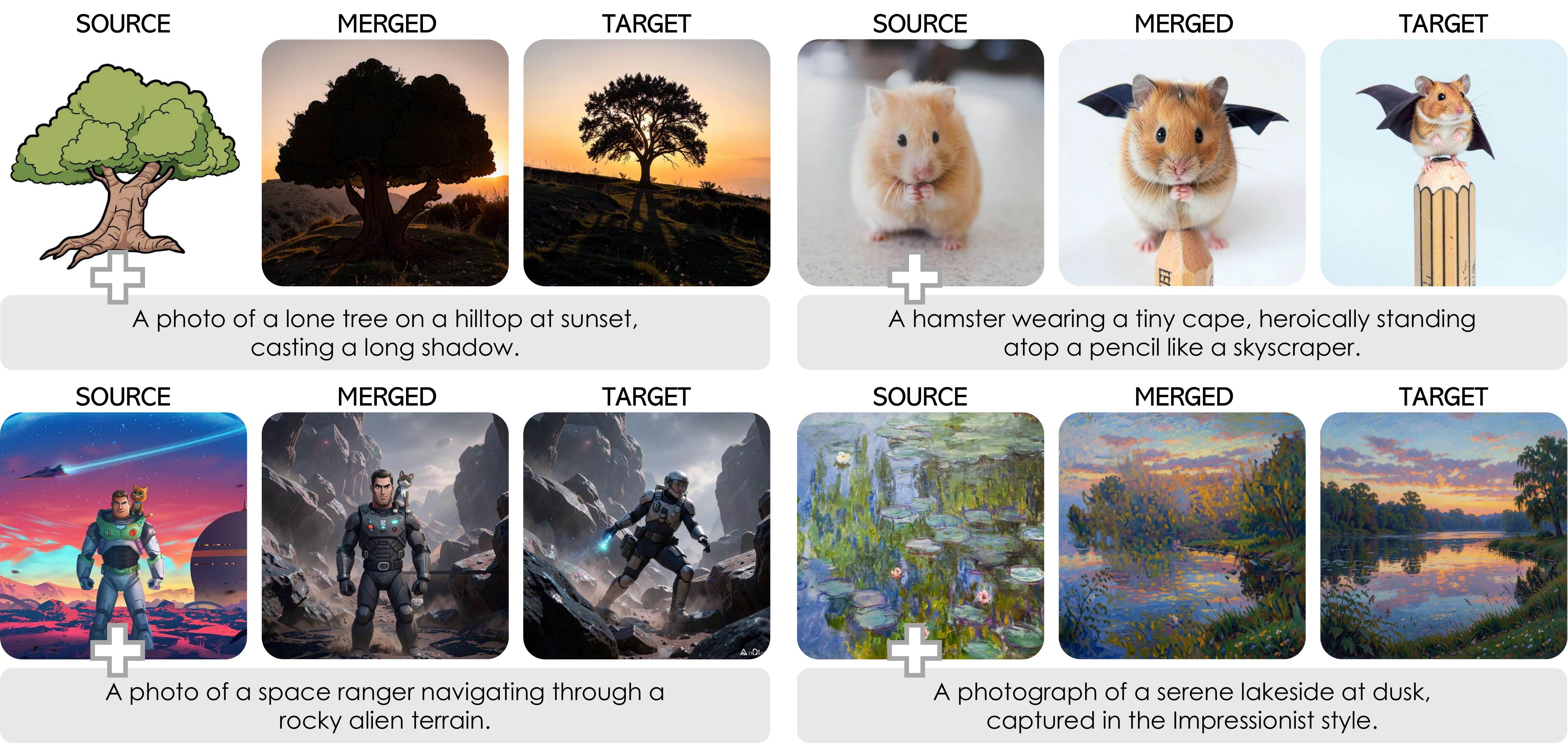}
    \vspace{-0.2cm}
    \caption{Qualitative examples of MAs-based image-conditioned transport on FLUX.2-klein.}
    \label{fig:img_prompt_klein}
    \vspace{-0.3cm}
\end{figure}

\begin{figure}[t]
    \centering    \includegraphics[width=0.98\linewidth]{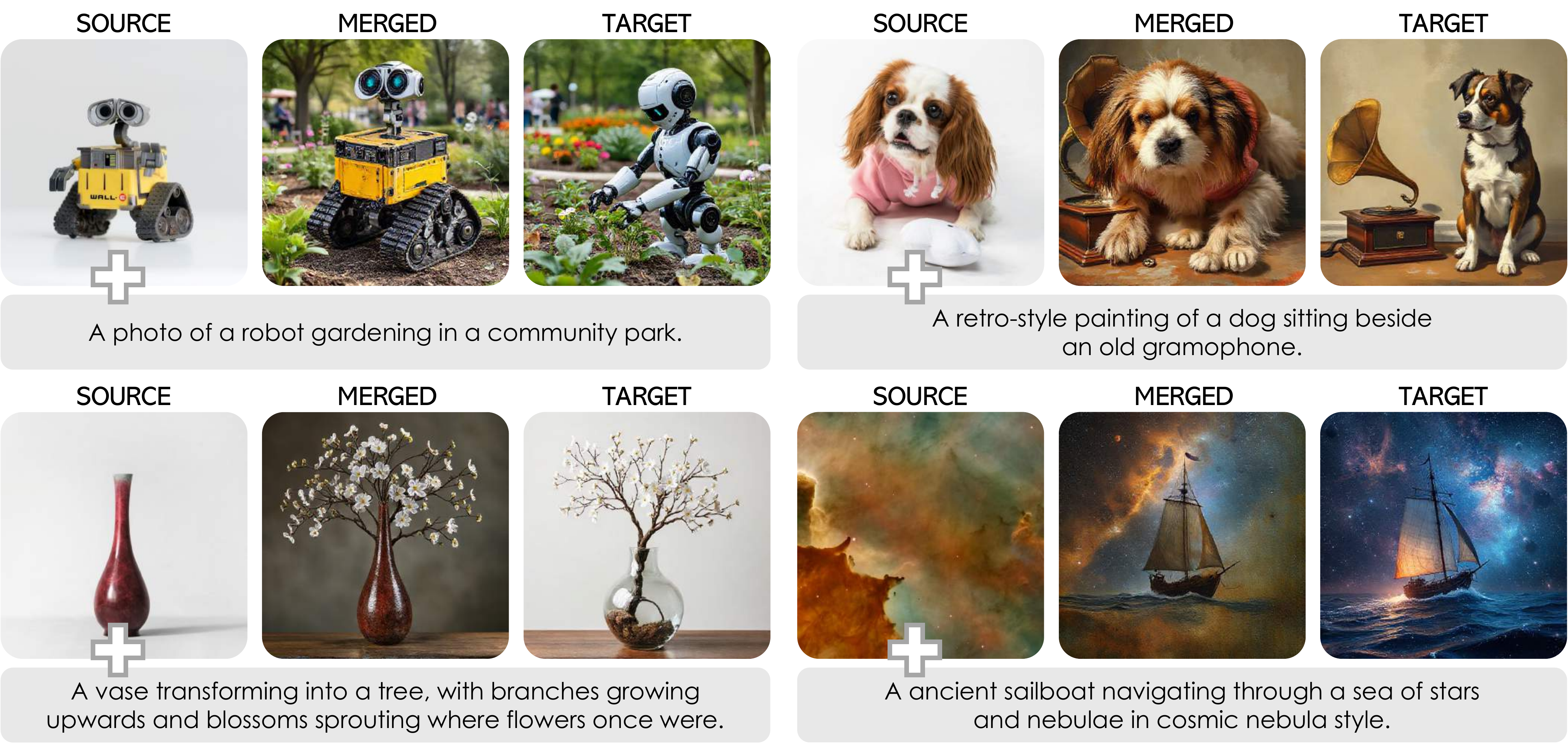}
    \vspace{-0.2cm}
    \caption{Qualitative examples of MAs-based image-conditioned transport on FLUX.1-schnell.}
    \label{fig:img_prompt_schnell}
    \vspace{-0.3cm}
\end{figure}

We provide additional qualitative results for prompt-to-prompt semantic transport across models in Figs.~\ref{fig:prompt_prompt_dev}-\ref{fig:prompt_prompt_schnell}. Each example shows a source, a target, and the merged output. Figs.~\ref{fig:img_prompt_dev}-\ref{fig:img_prompt_schnell} present image-conditioned semantic transport. Across both settings, the results exhibit consistent compositional behavior: salient elements from the source are integrated into the target scene while preserving its structure, lighting, and perspective. This is not a pixel-level mixture, the outputs remain visually coherent and semantically consistent, indicating that the transported information corresponds to high-level features. The effect is robust across multiple composition types, including object transfer, attribute transfer, and scene recomposition. In many cases, higher-level attributes such as artistic style, lighting, and semantic roles are preserved or reinterpreted. Notably, transferred subjects are not simply pasted into the scene, but are adapted to match the geometry, scale, and visual context of the target generation.

\section{Limitations and Societal Impacts}
\tit{Limitations} Our study is empirical in nature: we characterize the role of massive activations through interventional probes and demonstrate their use for semantic transport across generations, but we do not investigate the underlying reasons behind the emergence of this sparse subspace, nor what individual channels encode. Furthermore, the spatial masks induced by MAs reliably separate the salient subject from the background, but do not characterize what lies inside the salient region: they are agnostic to object identity and part-level structure. The transport mechanism inherits this granularity, transferring information within the salient region without distinguishing the entities it contains; finer, concept-aware masks would be required to disentangle and selectively manipulate individual semantic components. We leave both directions to future work. Finally, the transport mechanism operates on activations extracted from a specific generation instance, so artifacts and errors produced by the underlying generator on the source is inherited by the transported output.

 \tit{Societal Impacts} Our analysis operates on pretrained DiTs without additional training, improving interpretability and controllability of text-to-image generators at low compute cost. The image-conditioned transport mechanism shares the risks common to personalization methods (non-consensual imagery, copyright infringement, misleading content); since we build on pretrained backbones, we inherit their safety mechanisms without weakening them, and recommend pairing downstream deployments with standard mitigations such as prompt filtering and watermarking.


\clearpage
\begin{figure}[t]
    \centering    \includegraphics[width=0.98\linewidth]{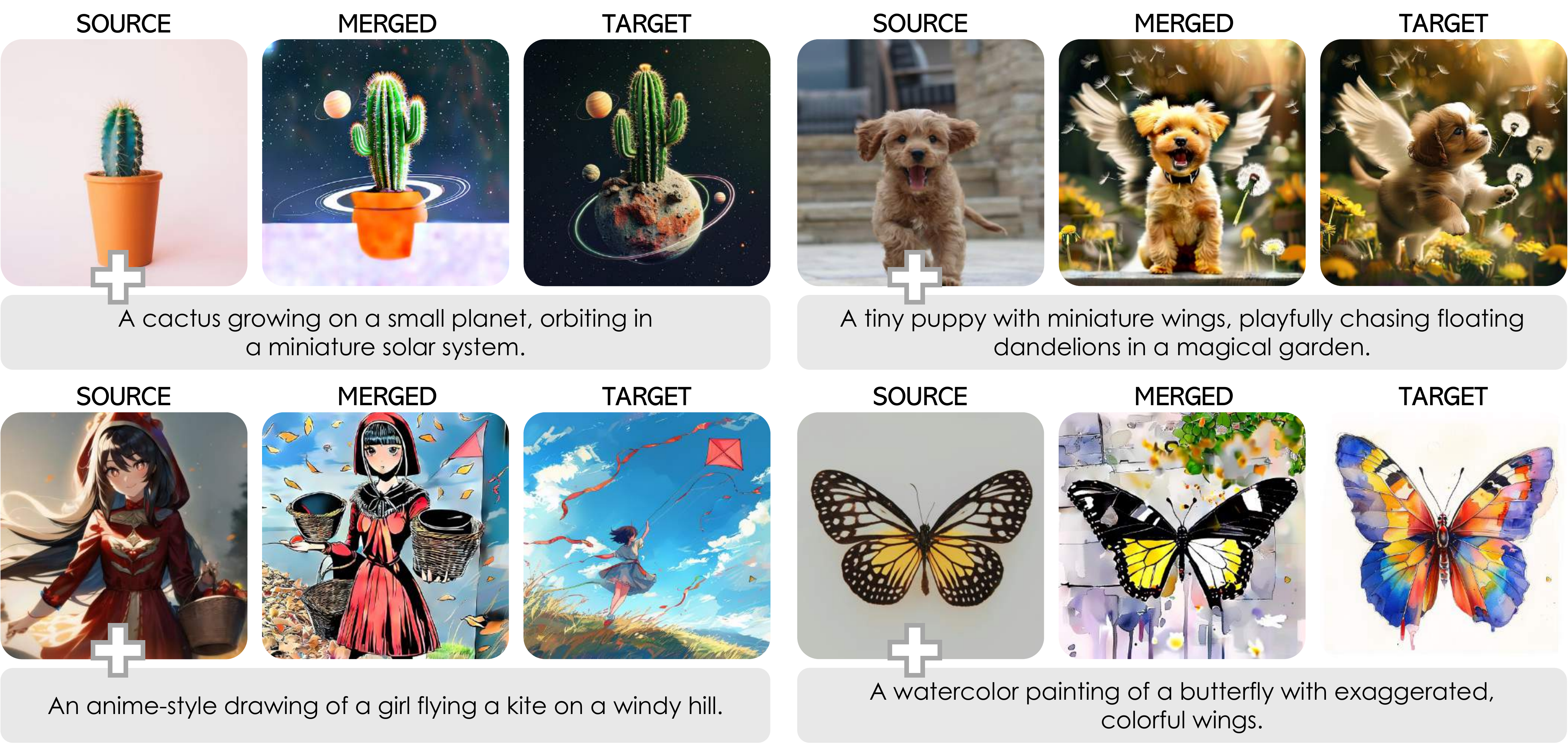}
    \vspace{-0.2cm}
    \caption{Qualitative examples of MAs-based image-conditioned transport on SANA1.5.}
    \label{fig:img_prompt_sana}
    \vspace{-0.3cm}
\end{figure}

\end{document}